\documentclass[journal]{IEEEtran}

\usepackage[amssymb]{SIunits}
\usepackage{graphicx}
\usepackage{amssymb}
\usepackage{amsthm}
\usepackage{multirow}
\usepackage{enumerate}
\usepackage{enumitem}
\usepackage{wrapfig}
\usepackage{seqsplit}

\usepackage{amsmath}
\usepackage[cmintegrals]{newtxmath}
\usepackage{bm} 
\usepackage{footnote}
\usepackage{threeparttable}
\usepackage{epsfig}
\usepackage{graphicx}
\usepackage{amsmath}
\usepackage{amssymb}
\usepackage{balance}
\usepackage{mathtools}
\usepackage{placeins}
\usepackage{pifont}
\usepackage[misc]{ifsym}
\usepackage[super]{nth}

\usepackage{url}            
\usepackage{booktabs}       
\usepackage{amsfonts}       
\usepackage{nicefrac}       
\usepackage{microtype}      
\usepackage{lipsum}
\usepackage[hidelinks,colorlinks=true,urlcolor=blue,citecolor=blue,allcolors=blue]{hyperref}
\usepackage{graphicx}
\usepackage{color}
\usepackage[table,dvipsnames]{xcolor}
\usepackage{todonotes}
\usepackage{marginnote}
\presetkeys{todonotes}{size=\tiny}{}

\usepackage{array}
\newcommand{\PreserveBackslash}[1]{\let\temp=\\#1\let\\=\temp}
\newcolumntype{C}[1]{>{\PreserveBackslash\centering}p{#1}}
\newcolumntype{R}[1]{>{\PreserveBackslash\raggedleft}p{#1}}
\newcolumntype{L}[1]{>{\PreserveBackslash\raggedright}p{#1}}

\newcommand{\rom}[1]{\uppercase\expandafter{\romannumeral #1\relax}}

\newcommand{\fref}[1]{Fig.~\ref{#1}}
\newcommand{\sref}[1]{Section~\ref{#1}}
\newcommand{\tref}[1]{Table~\ref{#1}}

\newcommand{\etal}{\textit{et al.}~}
\newcommand{\ie}{{i.e.},~}
\newcommand{\eg}{{e.g.},~}

\newcommand{\xx}{\ding{53}}
\newcommand{\cc}{\checkmark}

\graphicspath{%
    {./figures/}%
}

\ifCLASSOPTIONcompsoc
  \usepackage[caption=false,font=footnotesize,labelfont=sf,textfont=sf]{subfig}
\else
  \usepackage[caption=false,font=footnotesize]{subfig}
\fi

\ifCLASSOPTIONcompsoc
  \usepackage[nocompress]{cite}
\else
  \usepackage{cite}
\fi

\ifCLASSINFOpdf
\else
\fi
\usepackage{algorithm}
\usepackage{algorithmic}
\ifCLASSOPTIONcompsoc
 \usepackage[caption=false,font=normalsize,labelfont=sf,textfont=sf]{subfig}
\else
 \usepackage[caption=false,font=footnotesize]{subfig}
\fi
\usepackage{url}

\begin{document}
%
\title{AirSLAM: An Efficient and Illumination-Robust Point-Line Visual SLAM System}
%
%
%

\author{Kuan~Xu$^{1}$,~
        Yuefan Hao$^{2}$,~
        Shenghai Yuan$^{1}$,~\IEEEmembership{Member,~IEEE,}\\
        Chen~Wang$^{2}$,~\IEEEmembership{Senior Member,~IEEE,}
        Lihua~Xie$^{1}$,~\IEEEmembership{Fellow,~IEEE}
\thanks{This work is supported by the National Research Foundation of Singapore under its Medium-Sized Center for Advanced Robotics Technology Innovation.}
\thanks{$^{1}$Kuan Xu, Shenghai Yuan, and Lihua Xie are with the Centre for Advanced Robotics Technology Innovation (CARTIN), School of Electrical and Electronic Engineering, Nanyang Technological University, 50 Nanyang Avenue, Singapore 639798, {\tt\small \{kuan.xu,shyuan,elhxie\}@ntu.edu.sg}.}
\thanks{$^{2}$Yuefan Hao and Chen~Wang are with Spatial AI \& Robotics Lab, Computer Science and Engineering, University at Buffalo, Buffalo, NY 14260, USA {\tt\small yuefan.hao@outlook.com}, {\tt\small chenw@sairlab.org}.}}

%
%

\markboth{}%
{}
%



\maketitle

\begin{abstract}

In this paper, we present an efficient visual SLAM system designed to tackle both short-term and long-term illumination challenges. Our system adopts a hybrid approach that combines deep learning techniques for feature detection and matching with traditional backend optimization methods. Specifically, we propose a unified convolutional neural network (CNN) that simultaneously extracts keypoints and structural lines. These features are then associated, matched, triangulated, and optimized in a coupled manner. Additionally, we introduce a lightweight relocalization pipeline that reuses the built map, where keypoints, lines, and a structure graph are used to match the query frame with the map. To enhance the applicability of the proposed system to real-world robots, we deploy and accelerate the feature detection and matching networks using C++ and NVIDIA TensorRT. Extensive experiments conducted on various datasets demonstrate that our system outperforms other state-of-the-art visual SLAM systems in illumination-challenging environments. Efficiency evaluations show that our system can run at a rate of $73\hertz$ on a PC and $40\hertz$ on an embedded platform. Our implementation is open-sourced: {\url{https://github.com/sair-lab/AirSLAM}}.

\end{abstract}


%
\IEEEpeerreviewmaketitle

\section{Introduction}

\IEEEPARstart{V}{isual} simultaneous localization and mapping (vSLAM) is essential for robot navigation due to its favorable balance between cost and accuracy \cite{macario2022comprehensive}. 
Compared to LiDAR SLAM, vSLAM utilizes more cost-effective and compact sensors to achieve accurate localization, thus broadening its range of potential applications \cite{yuan2021survey}. Moreover, cameras can capture richer and more detailed information, which enhances their potential for providing robust localization.

Despite the recent advancements, the present vSLAM systems still struggle with severe lighting conditions \cite{mur2015orb,qin2018vins,gomez2019pl,cadena2016past}, which can be summarized into two categories.
First, feature detection and tracking often fail due to drastic changes or low light, severely affecting the quality of the estimated trajectory \cite{zuniga2020vi, savinykh2022darkslam}. 
Second, when the visual map is reused for relocalization, lighting variations could significantly reduce the success rate \cite{sarlin2019coarse, toft2020long}. In this paper, we refer to the first issue as the short-term illumination challenge, which impacts pose estimation between two temporally adjacent frames, and the second as the long-term illumination challenge, which affects matching between the query frame and an existing map.

Present methods usually focus on only one of the above challenges.
For example, various image enhancement \cite{gu2021drms, yu2022afe, gomez2018learning} and image normalization algorithms \cite{scandaroli2012improving, usenko2019visual} have been developed to ensure robust tracking. These methods primarily focus on maintaining either global or local brightness consistency, yet they often fall short of handling all types of challenging lighting conditions \cite{park2017illumination}. 
Some systems have addressed this issue by training a VO or SLAM network on large datasets containing diverse lighting conditions \cite{wang2021tartanvo, teed2021droid, fu2024islam}. However, they have difficulty producing a map suitable for long-term localization.
Some methods can provide illumination-robust relocalization, but they usually require map building under good lighting conditions \cite{labbe2022multi, labbe2019rtab}.
In real-world robot applications, these two challenges often arise simultaneously, necessitating a unified system capable of addressing both.

Furthermore, many of the aforementioned systems incorporate intricate neural networks, relying on powerful GPUs to run in real-time. They lack the efficiency necessary for deployment on resource-constrained platforms, such as warehouse robots. 
These limitations impede the transition of vSLAM from laboratory research to industrial applications.

In response to these gaps, this paper introduces AirSLAM.
Observing that line features can improve the accuracy and robustness of vSLAM systems \cite{gomez2019pl, xu2023airvo, kannapiran2023stereo}, we integrate both point and line features for tracking, mapping, optimization, and relocalization. To achieve a balance between efficiency and performance, we design our system as a hybrid system, employing learning-based methods for feature detection and matching, and traditional geometric approaches for pose and map optimization. Additionally, to enhance the efficiency of feature detection, we developed a unified model capable of simultaneously detecting point and line features. We also address long-term localization challenges by proposing a multi-stage relocalization strategy, which effectively reuses our point-line map.
In summary, our contributions include
\begin{itemize}
  \item We propose a novel point-line-based vSLAM system that combines the efficiency of traditional optimization techniques with the robustness of learning-based methods. Our system is resilient to both short-term and long-term illumination challenges while remaining efficient enough for deployment on embedded platforms.
  \item We develop a unified model for both keypoint and line detection, which we call PLNet. To our knowledge, PLNet is the first model capable of simultaneously detecting both point and line features. Furthermore, we associate these two types of features and jointly utilize them for tracking, mapping, and relocalization tasks.
  \item We propose a multi-stage relocalization method based on both point and line features, utilizing both appearance and geometry information. This method can provide fast and illumination-robust localization in an existing visual map using only a single image. 
  \item We conduct extensive experiments to demonstrate the efficiency and effectiveness of the proposed methods. The results show that our system achieves accurate and robust mapping and relocalization performance under various illumination-challenging conditions. Additionally, our system is also very efficient. It runs at a rate of $73\hertz$ on a PC and $40\hertz$ on an embedded platform.
\end{itemize}
In addition, our engineering contributions include deploying and accelerating feature detection and matching networks using C++ and NVIDIA TensorRT, facilitating their deployment on real robots. We release all the source code at \textcolor{blue}{\url{https://github.com/sair-lab/AirSLAM}} to benefit the community.

This paper extends our conference paper, AirVO \cite{xu2023airvo}. AirVO utilizes SuperPoint \cite{detone2018superpoint} and LSD \cite{von2012lsd} for feature detection, and SuperGlue \cite{sarlin2020superglue} for feature matching. It achieves remarkable performance in environments with changing illumination. However, as a visual-only odometry, it primarily addresses short-term illumination challenges and cannot reuse a map for drift-free relocalization. 
Additionally, despite carefully designed post-processing operations, the modified LSD is still not stable enough for long-term localization. It relies on image gradient information rather than environmental structural information, rendering it susceptible to varying lighting conditions.
In this version, we introduce substantial improvements, including:
\begin{itemize}
    \item We design a unified CNN to detect both point and line features, enhancing the stability of feature detection in illumination-challenging environments. Additionally, the more efficient LightGlue \cite{lindenberger2023lightglue} is used for feature matching.
    \item We extend our system to support both stereo-only and stereo-inertial data, increasing its reliability when an inertial measurement unit (IMU) is available.
    \item We incorporate loop closure detection and map optimization, forming a complete vSLAM system.
    \item We design a multi-stage relocalization module based on both point and line features, enabling our system to effectively handle long-term illumination challenges.
\end{itemize}
The remainder of this article is organized as follows. In \sref{sec:related-work}, we discuss the relevant literature. \sref{sec:system_architecture} presents an overview of the complete system pipeline. The proposed PLNet is presented in \sref{sec:feature_detection}. In \sref{sec:visual_odometry}, we introduce the visual-inertial odometry based on PLNet. \sref{sec:optimization_relocalization} shows how to optimize the map offline and reuse it online. The detailed experimental results are presented in \sref{sec:experiments} to verify the efficiency, accuracy, and robustness of AirSLAM. This article is concluded in \sref{sec:conclusion}.

\section{Related work}\label{sec:related-work}

\subsection{Keypoint and Line Detection for vSLAM}\label{sec:related-ground}
\subsubsection{Keypoint Detection}
Various handcrafted keypoint features e.g., ORB \cite{rublee2011orb}, FAST \cite{viswanathan2009features}, and BRISK \cite{leutenegger2011brisk}, have been proposed and applied to VO and vSLAM systems. They are usually efficient but not robust enough in challenging environments \cite{sarlin2020superglue, sarlin2019coarse}.
With the development of deep learning techniques, more and more learning-based features are proposed and used to replace the handcrafted features in vSLAM systems. Rong \etal \cite{kang2019df} introduce TFeat network \cite{balntas2016learning} to extract descriptors for FAST corners and apply it to a traditional vSLAM pipeline. Tang \etal \cite{tang2019gcnv2} use a neural network to extract robust keypoints and binary feature descriptors with the same shape as the ORB. Han \etal \cite{han2020superpointvo} combine SuperPoint \cite{detone2018superpoint} feature extractor with a traditional back-end. Bruno \etal proposed LIFT-SLAM \cite{bruno2021lift}, where they use LIFT \cite{yi2016lift} to extract features. Li \etal \cite{li2020dxslam} replace the ORB feature with SuperPoint in ORB-SLAM2 and optimize the feature extraction with the Intel OpenVINO toolkit. Zhan \etal \cite{zhan2024imatching} proposed a new self-supervised training scheme for learning-based features, using bundle adjustment and bi-level optimization as a supervision signal. Some other learning-based features, e.g., R2D2 \cite{revaud2019r2d2} and DISK \cite{tyszkiewicz2020disk}, and D2Net \cite{dusmanu2019d2}, are also being attempted to be applied to vSLAM systems, although they are not yet efficient enough \cite{hu2021joint, anagnostopoulos2023reviewing}.

\subsubsection{Line Detection}
Currently, most point-line-based vSLAM systems use the LSD \cite{von2012lsd} or EDLines \cite{akinlar2011edlines} to detect line features because of their good efficiency \cite{gomez2019pl, he2018pl, zhou2022edplvo, zou2019structvio, chen2024vpl}. Although many learning-based line detection methods, e.g., SOLD2 \cite{Pautrat_Lin_2021_CVPR}, AirLine \cite{lin2023airline}, and HAWP \cite{xue2023holistically}, have been proposed and shown better robustness in challenging environments, they are difficult to apply to real-time vSLAM systems due to lacking efficiency. For example, Kannapiran \etal propose StereoVO \cite{kannapiran2023stereo}, where they choose SuperPoint \cite{detone2018superpoint} and SOLD2 \cite{Pautrat_Lin_2021_CVPR} to detect keypoints and line segments, respectively. Despite achieving good performance in dynamic lighting conditions, StereoVO can only run at a rate of about $7\hertz$ on a good GPU.

\vspace{-5pt}
\subsection{Short-Term Illumination Challenge}
Several handcrafted methods have been proposed to improve the robustness of VO and vSLAM to challenging illumination. 
DSO \cite{engel2017direct} models brightness changes and jointly optimizes camera poses and photometric parameters. 
DRMS \cite{gu2021drms} and AFE-ORB-SLAM \cite{yu2022afe} utilize various image enhancements. 
Some systems try different methods, such as ZNCC, the locally-scaled sum of squared differences (LSSD), and dense descriptor computation, to achieve robust tracking \cite{scandaroli2012improving, crivellaro2014robust, usenko2019visual}. 
These methods mainly focus on either global or local illumination change for all kinds of images, however, lighting conditions often affect the scene differently in different areas \cite{park2017illumination}.
Other related methods include that of Huang and Liu \cite{huang2019robust}, which presents a multi-feature extraction algorithm to extract two kinds of image features when a single-feature algorithm fails to extract enough feature points. Kim \etal \cite{kim2019autonomous} employ a patch-based affine illumination model during direct motion estimation. Chen \etal \cite{chen2021robust} minimize the normalized information distance with nonlinear least square optimization for image registration. Alismail \etal \cite{alismail2016direct} propose a binary feature descriptor using a descriptor assumption to avoid brightness constancy.

Compared with handcrafted methods, learning-based methods have shown better performance. Savinykh \etal \cite{savinykh2022darkslam} propose DarkSLAM, where Generative Adversarial Network (GAN) \cite{creswell2018generative} is used to enhance input images. Pratap Singh \etal \cite{pratap2023twilight} compare different learning-based image enhancement methods for vSLAM in low-light environments. TartanVO \cite{wang2021tartanvo}, DROID-SLAM \cite{teed2021droid}, and iSLAM \cite{fu2024islam} train their VO or SLAM networks on the TartanAir dataset \cite{wang2020tartanair}, which is a large simulation dataset that contains various lighting conditions, therefore, they are very robust in challenging environments. However, they usually require good GPUs and long training times. Besides, DROID-SLAM runs very slowly and is difficult to apply to real-time applications on resource-constrained platforms. TartanVO and iSLAM are more efficient, but they cannot achieve performance as accurately as traditional vSLAM systems.

\vspace{-5pt}
\subsection{Long-Term Illumination Challenge} \label{sec:related-correlation}
Currently, most SLAM systems still use the bag of words (BoW) \cite{galvez2012bags} for loop closure detection and relocalization due to its good balance between efficiency and effectiveness \cite{campos2021orb, qin2018vins, rosinol2020kimera}. To make the relocalization more robust to large illumination variations, Labb{\'e} \etal \cite{labbe2022multi} propose the multi-session relocalization method, where they combine multiple maps generated at different times and in various illumination conditions. 
DXSLAM \cite{li2020dxslam} uses NetVLAD \cite{arandjelovic2016netvlad} for the coarse
image retrieval and SuperPoint with a binary descriptor for keypoint matching between the query frame and candidates.

Another similar task in the robotics and computer vision communities is the visual place recognition (VPR) problem, where many researchers handle the localization problem with image retrieval methods \cite{arandjelovic2016netvlad, keetha2023anyloc}. These VPR solutions try to find images most similar to the query image from a database. They usually cannot directly provide accurate pose estimation which is needed in robot applications. Sarlin \etal address this and propose Hloc \cite{sarlin2019coarse}. They use a global retrieval to obtain several candidates and match local features within those candidates. The Hloc toolbox has integrated many image retrieval methods, local feature extractors, and matching methods, and it is currently the SOTA system. Yan \etal \cite{yan2023long} propose a long-term visual localization method for mobile platforms, however, they rely on other sensors, e.g., GPS, compass, and gravity sensor, for the coarse location retrieval.

\begin{figure}[t]
    \vspace{0.5em}
    \centering
    \includegraphics[width=0.985\linewidth]{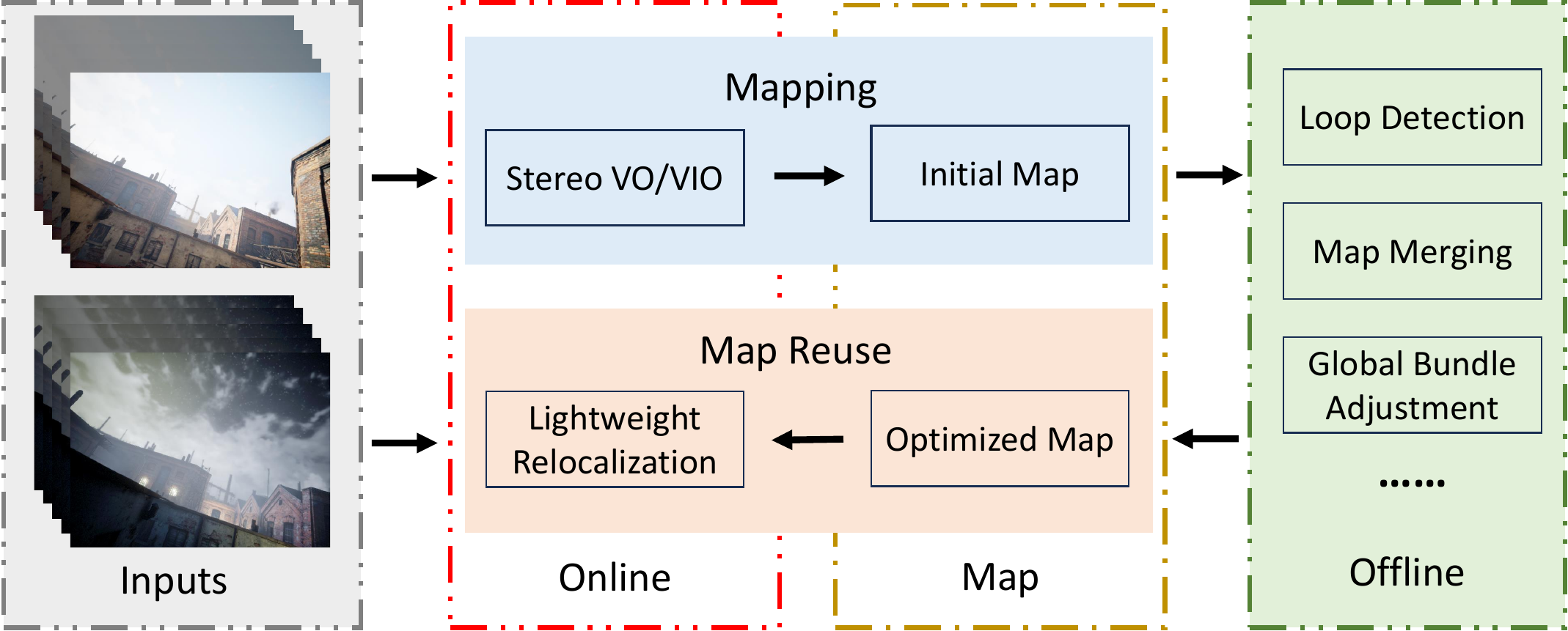}
    \caption{The proposed system consists of three main parts: online stereo VO/VIO, offline map optimization, and online relocalization. The VO/VIO module uses the mapping image sequences to build an initial map. Then the initial map is processed offline and an optimized map is outputted. The optimized map can be used for the one-shot relocalization.}
    \label{fig:system_arch}
    \vspace{-1.em}
\end{figure}

\section{System Overview}\label{sec:overview}

\subsection{Notations} \label{sec:so_notation}
In this paper, $\mathbb{R}$ represents the set of real numbers, and $\mathbb{R}^m$ denotes the $m$-dimensional real vector space. The transpose of a vector or matrix is written as $(\cdot)^\top$. For a vector $\mathbf{x} \in \mathbb{R}^m$, $\lVert \mathbf{x} \rVert$ represents its Euclidean norm, while $\lVert \mathbf{x} \rVert^2_{\Sigma}$ is shorthand for $\mathbf{x}^\top \Sigma \mathbf{x}$. The transformation, rotation, and translation from the $b$-coordinate system to the $a$-coordinate system are denoted by $\mathbf{T}_{ab} \in \text{SE(3)}$, $\mathbf{R}_{ab} \in \text{SO(3)}$, and $\mathbf{t}_{ab} \in \mathbb{R}^3$, respectively. We use $(\cdot)_c$ to indicate a vector in the camera frame and $(\cdot)_w$ to indicate a vector in the global world frame. Throughout the paper, super/subscripts may be omitted for brevity, as long as the meaning remains unambiguous within the given context.

\begin{figure}[t]
    \vspace{0.4em}
    \centering
    \setlength{\abovecaptionskip}{-0.03cm}
    \includegraphics[width=0.93\linewidth]{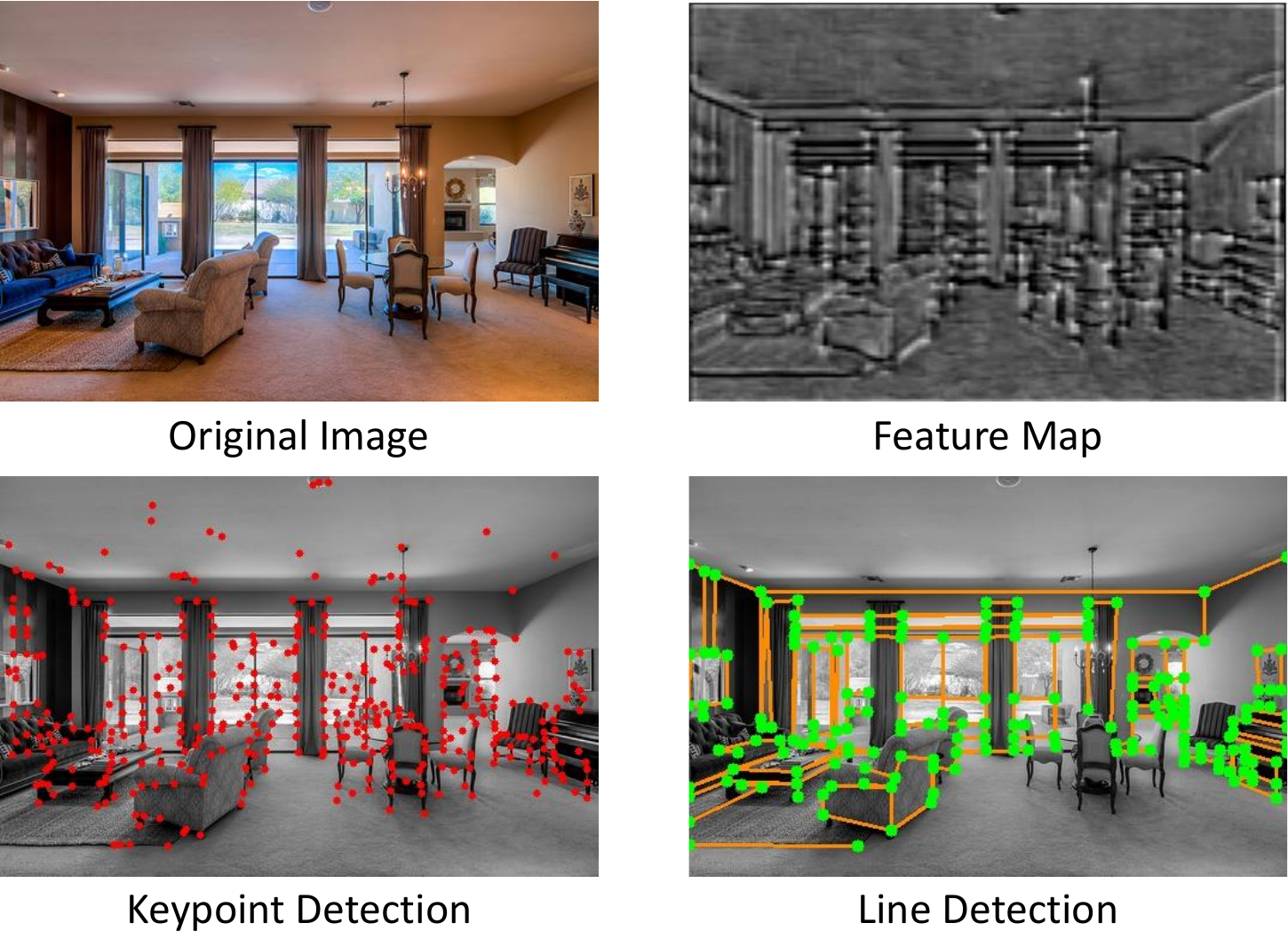}
    \caption{We visualize the feature map (top right) and detected keypoints (bottom left) of a keypoint detection model, and the detected structural lines (bottom right) of a line detection model. The overlap of keypoints and junctions, and the edge information in the feature map inspire the design of our PLNet.  }
    \label{fig:feature_observation}
    \vspace{-1.0em}
\end{figure}

\subsection{System Architecture} \label{sec:system_architecture}

We believe that a practical vSLAM system should possess the following features:
\begin{itemize}
\item High efficiency. The system should have real-time performance on resource-constrained platforms.  
\item Scalability. The system should be easily extensible for various purposes and real-world applications.
\item Easy to deploy. The system should be easy to deploy on real robots and capable of achieving robust localization.
\end{itemize}

Therefore, we design a system as shown in \fref{fig:system_arch}. The proposed system is a hybrid system as we need the robustness of data-driven approaches and the accuracy of geometric methods. It consists of three main components: stereo VO/VIO, offline map optimization, and lightweight relocalization. (1) Stereo VO/VIO: We propose a point-line-based visual odometry that can handle both stereo and stereo-inertial inputs. (2) Offline map optimization: We implement several commonly used plugins, such as loop detection, pose graph optimization, and global bundle adjustment. The system is easily extensible for other map-processing purposes by adding customized plugins. For example, we have implemented a plugin to train a scene-dependent junction vocabulary using the endpoints of line features, which is utilized in our lightweight multi-stage relocalization. (3) Lightweight relocalization: We propose a multi-stage relocalization method that improves efficiency while maintaining effectiveness. In the first stage, keypoints and line features are detected using the proposed PLNet, and several candidates are retrieved using a keypoint vocabulary trained on a large dataset. In the second stage, most false candidates are quickly filtered out using a scene-dependent junction vocabulary and a structure graph. In the third stage, feature matching is performed between the query frame and the remaining candidates to find the best match and estimate the pose of the query frame. Since feature matching in the third stage is typically time-consuming, the filtering process in the second stage enhances the efficiency of our system compared to other two-stage relocalization systems.

We transfer some time-consuming processes, e.g., loop closure detection, pose graph optimization, and global bundle adjustment, to the offline stage. This improves the efficiency of our online mapping module. In many practical applications, such as warehouse robotics, a map is typically built by one robot and then reused by others. Our system is designed with these applications in mind. The lightweight mapping and map reuse modules can be easily deployed on resource-constrained robots, while the offline optimization module can run on a more powerful computer for various map manipulations, such as map editing and visualization. The mapping robot uploads the initial map to the computer, which then distributes the optimized map to other robots, ensuring drift-free relocalization.
In the following sections, we introduce our feature detection and visual odometry (VO) pipeline in \sref{sec:feature_detection} and \sref{sec:visual_odometry}, respectively. The offline optimization and relocalization modules are presented in \sref{sec:optimization_relocalization}.

\begin{figure}[t]
    \vspace{0.5em}
    \centering
    \setlength{\abovecaptionskip}{-0.3cm}
    \includegraphics[width=0.99\linewidth]{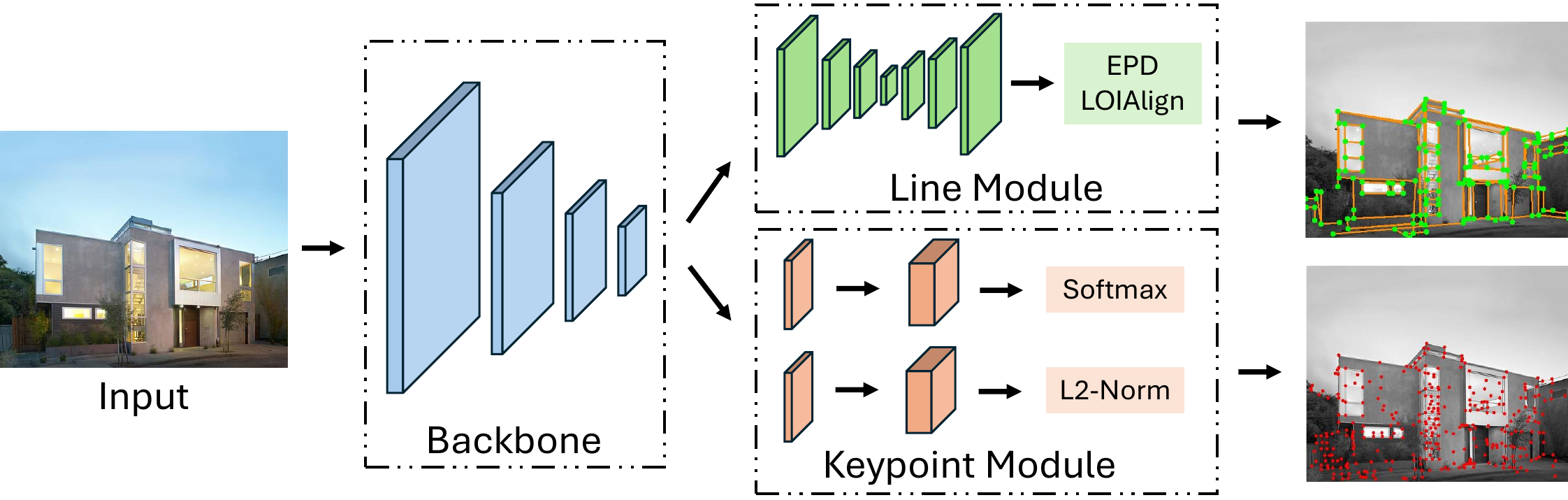}
    \caption{The framework of the proposed PLNet. It consists of the shared backbone, the keypoint module, and the line module.}
    \label{fig:plnet}
    \vspace{-1.em}
\end{figure}

\section{Feature Detection}\label{sec:feature_detection}

\subsection{Motivation} \label{sec:fd_motivation}

With advancements in deep learning technology, learning-based feature detection methods have demonstrated more stable performance in illumination-challenging environments compared to traditional methods. However, existing point-line-based VO/VIO and SLAM systems typically detect keypoints and line features separately. While it is acceptable for handcrafted methods due to their efficiency, the simultaneous application of keypoint detection and line detection networks in VO/VIO or SLAM systems, especially in stereo configurations, often hinders real-time performance on resource-constrained platforms. Consequently, we aim to design an efficient unified model that can detect keypoints and line features concurrently.

However, achieving a unified model for keypoint and line detection is challenging, as these tasks typically require different real-image datasets and training procedures. Keypoint detection models are generally trained on large datasets comprising diverse images and depend on either a boosting step or the correspondences of image pairs for training \cite{detone2018superpoint, tyszkiewicz2020disk, revaud2019r2d2}. For line detection, we find wireframe parsing methods \cite{zhou2019end, xue2023holistically} can provide stronger geometric cues than the self-supervised models \cite{Pautrat_Lin_2021_CVPR, pautrat2023deeplsd} as they are able to detect longer and more complete lines. The wireframe includes all prominent straight lines and their junctions within the scene, providing an efficient and accurate representation of large-scale geometry and object shapes \cite{huang2018learning}. However, these methods are trained on the Wireframe dataset \cite{huang2018learning}, which is limited in size with only 5,462 discontinuous images. In the following sections, we will address this challenge and demonstrate how to train a unified model capable of performing both tasks. It is important to note that in this paper, the term ``line detection" refers specifically to the wireframe parsing task.

\begin{figure}[t]
    \vspace{0.5em}
    \centering
    \setlength{\abovecaptionskip}{-0.03cm}
    \includegraphics[width=0.95\linewidth]{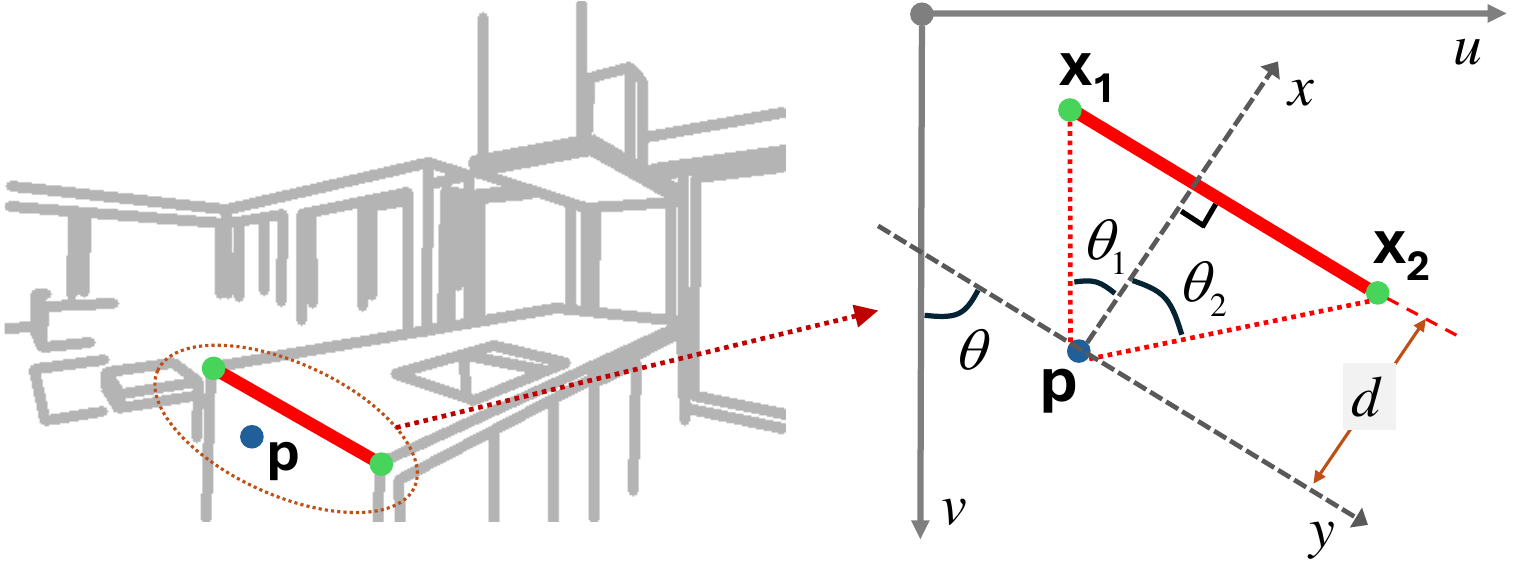}
    \caption{We use four parameters, $d, \theta, \theta_1, \theta_2$, to encode a line into a point $\mathbf{p}$ within the attraction region field.}
    \label{fig:line_enconding}
    \vspace{-0.5em}
\end{figure}

\subsection{Architecture Design} \label{sec:fd_architecture}
As shown in \fref{fig:feature_observation}, we have two findings when visualizing the results of the keypoint and line detection networks: (1) Most junctions (endpoints of lines) detected by the line detection model are also selected as keypoints by the keypoint detection model. (2) The feature maps outputted by the keypoint detection model contain the edge information. Therefore, we argue that a line detection model can be built on the backbone of a pre-trained keypoint detection model. 
Based on this assumption, we design the PLNet to detect keypoints and lines in a unified framework. As shown in \fref{fig:plnet}, it consists of the shared backbone, the keypoint module, and the line module. 

\textbf{Backbone:}
We follow SuperPoint \cite{detone2018superpoint} to design the backbone for its good efficiency and effectiveness. It uses 8 convolutional layers and 3 max-pooling layers. The input is the grayscale image sized $H \times W$. The outputs are $H \times W \times 64$, $\frac{H}{2} \times \frac{W}{2} \times 64$, $\frac{H}{4} \times \frac{W}{4} \times 128$, $\frac{H}{8} \times \frac{W}{8} \times 128$ feature maps.

\textbf{Keypoint Module:} 
We also follow SuperPoint \cite{detone2018superpoint} to design the keypoint detection header. It has two branches: the score branch and the descriptor branch. The inputs are $\frac{H}{8} \times \frac{W}{8} \times 128$ feature maps outputted by the backbone. The score branch outputs a tensor sized $\frac{H}{8} \times \frac{W}{8} \times 65$. The 65 channels correspond to an $8 \times 8$ grid region and a dustbin indicating no keypoint. The tensor is processed by a softmax and then resized to $H \times W$. The descriptor branch outputs a tensor sized $\frac{H}{8} \times \frac{W}{8} \times 256$, which is used for interpolation to compute descriptors of keypoints.

\textbf{Line Module:} 
This module takes feature maps from the backbone as inputs.
It consists of a U-Net-like CNN and the line detection header. We modify the U-Net \cite{ronneberger2015u} to make it contain fewer convolutional layers and thus be more efficient. The U-Net-like CNN is to increase the receptive field as detecting lines requires a larger receptive field than detecting keypoints.
The EPD LOIAlign \cite{xue2023holistically} is used to process the outputs of the line module and finally outputs junctions and lines.

\begin{figure}[t]
    \vspace{0.5em}
    \centering
    \setlength{\abovecaptionskip}{-0.03cm}
    \includegraphics[width=0.95\linewidth]{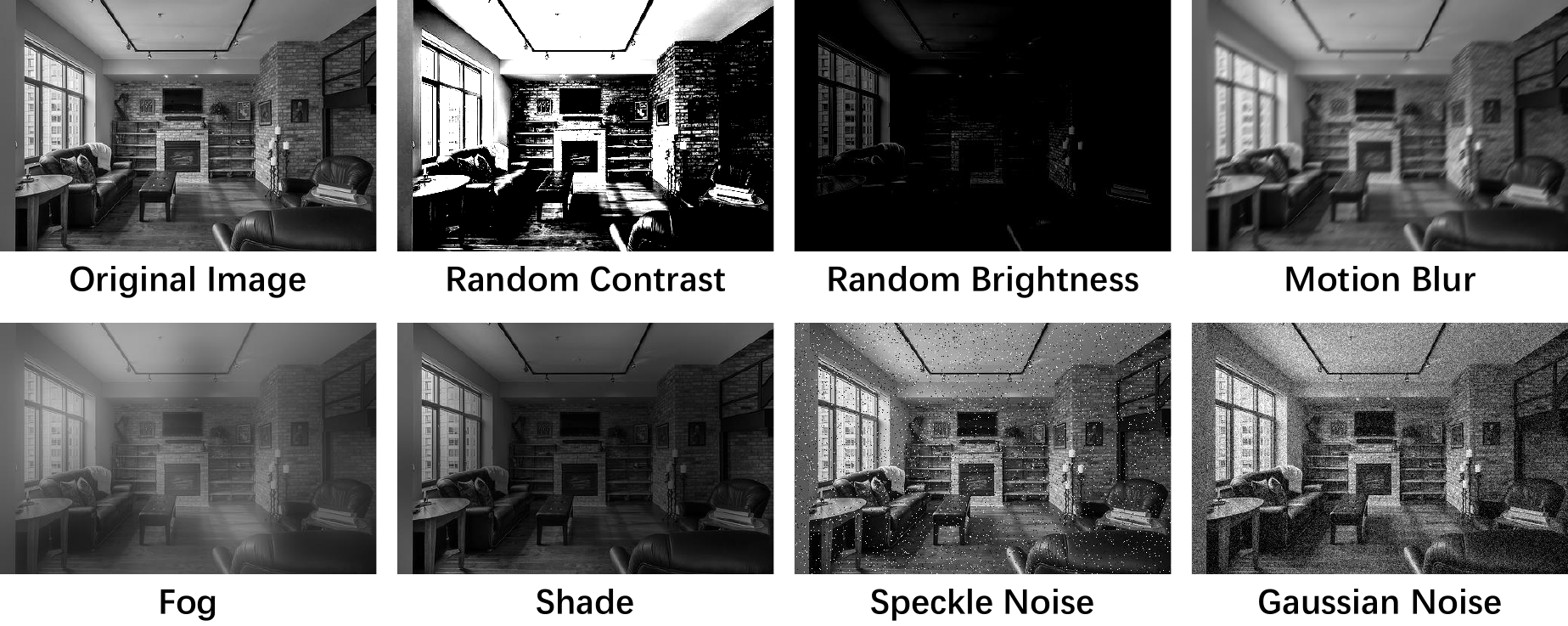}
    \caption{We use seven types of photometric data augmentation to train our PLNet to make it more robust to challenging illumination.}
    \label{fig:data_aug}
    \vspace{-0.5em}
\end{figure}

\subsection{Network Training}
Due to the training problem described in \sref{sec:fd_motivation} and the assumption in \sref{sec:fd_architecture}, we train our PLNet in two rounds. In the first round, only the backbone and the keypoint detection module are trained, which means we need to train a keypoint detection network. In the second round, the backbone and the keypoint detection module are fixed, and we only train the line detection module on the Wireframe dataset. 
We skip the details of the first round as they are very similar to \cite{detone2018superpoint}. Instead, we present the training of the line detection module.

\begin{figure*}[t]
    \vspace{0.5em}
    \centering
    \setlength{\abovecaptionskip}{-0.05cm}
    \includegraphics[width=0.95\linewidth]{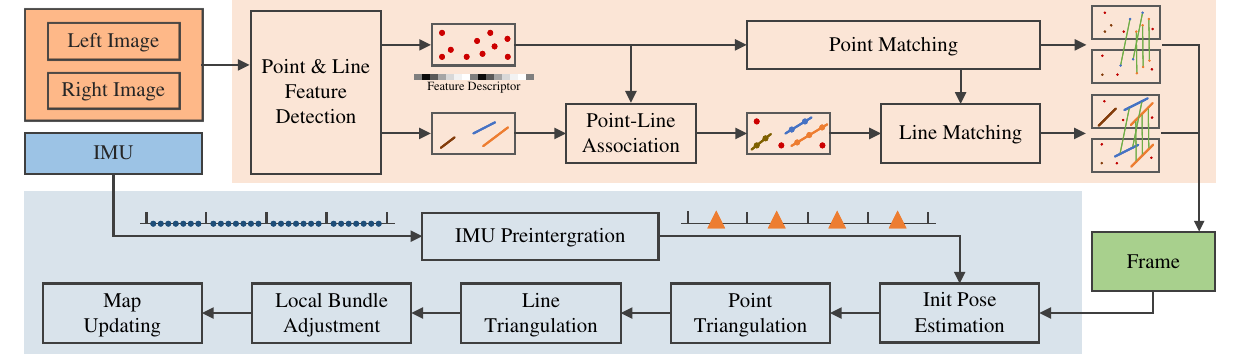}
    \caption{The framework of our visual(-inertial) odometry. The system is split into two main threads, which are represented by two different colored regions. Note that the IMU input is not strictly required. The system is optional to use stereo data or stereo-inertial data.
    }
    \label{fig:pipeline}
    \vspace{-1.0em}
\end{figure*}

\textbf{Line Encoding:} 
We adopt the attraction region field \cite{xue2023holistically} to encode line segments. As shown in \fref{fig:line_enconding}, for a line segment $\mathbf{l} = (\mathbf{x_1}, \mathbf{x_2})$, where $\mathbf{x_1}$ and $\mathbf{x_2}$ are two endpoints of $\mathbf{l}$, and a point $\mathbf{p}$ in the attraction region of $\mathbf{l}$, four parameters and $\mathbf{p}$ are used to encode $\mathbf{l}$:
\begin{equation}\label{eq:line_encoder}
   \mathbf{p} \left(\mathbf{l} \right) = \left(d, \theta, \theta_1, \theta_2 \right),
\end{equation}
where $d$ is the distance from $\mathbf{p}$ to $\mathbf{l}$, $\theta$ is the angle between $\mathbf{l}$ and the $\mathbf{v}$-axis of the image, $\theta_1$ is the angle between $\mathbf{px_1}$ and the perpendicular line from $\mathbf{p}$ to $\mathbf{l}$, and $\theta_2$ is the angle between $\mathbf{px_2}$ and the perpendicular line. The network can predict these four parameters for point $\mathbf{p}$ and then  
$\mathbf{l}$ can be decoded through:
\begin{equation}\label{eq:line_decoder}
   \mathbf{l} = d \cdot \left[\begin{array}{cc} \cos{\theta} & -\sin{\theta} \\ \sin{\theta} & \cos{\theta} \end{array}\right] \left[\begin{array}{cc} 1 & 1 \\ \tan{\theta_1} & \tan{\theta_2} \end{array}\right] + \left[\begin{array}{cc} \mathbf{p} & \mathbf{p} \end{array}\right].
\end{equation}

\textbf{Line Prediction:} 
The line detection module outputs a tensor sized $\frac{H}{4} \times \frac{W}{4} \times 4$ to predict parameters in \eqref{eq:line_encoder} and a heatmap to predict junctions. For each decoded line segment by \eqref{eq:line_decoder}, two junctions closest to its endpoints will be selected to form a line proposal with it. Proposals with the same junctions will be deduplicated and only one is retained. Then the EPD LOIAlign \cite{xue2023holistically} and a head classifier are applied to decide whether the line proposal is a true line feature.

\textbf{Line Module Training:} 
We use the $L1$ loss to supervise the prediction of parameters in \eqref{eq:line_encoder} and the binary cross-entropy loss to supervise the junction heatmap and the head classifier. The total loss is the sum
of them.  As shown in \fref{fig:data_aug}, to improve the robustness of line detection in illumination-challenging environments, seven types of photometric data augmentation are applied to process training images.
The training uses the ADAM optimizer \cite{kingma2014adam} with the learning rate $lr = 4e\text{-}4$ in the first 35 epochs and $lr = 4e\text{-}5$ in the last 5 epochs.

\section{Stereo Visual Odometry}\label{sec:visual_odometry}

\subsection{Overview}  \label{sec:vo_overview}

The proposed point-line-based stereo visual odometry is shown in \fref{fig:pipeline}. It is a hybrid VO system utilizing both the learning-based front-end and the traditional optimization backend. For each stereo image pair, we first employ the proposed PLNet to extract keypoints and line features. Then a GNN (LightGlue \cite{lindenberger2023lightglue}) is used to match keypoints. In parallel, we associate line features with keypoints and match them using the keypoint matching results. After that, we perform an initial pose estimation and reject outliers. Based on the results, we triangulate the 2D features of keyframes and insert them into the map. Finally, the local bundle adjustment will be performed to optimize points, lines, and keyframe poses. In the meantime, if an IMU is accessible, its measurements will be processed using the IMU preintegration method \cite{forster2016manifold}, and added to the initial pose estimation and local bundle adjustment.

Applying both learning-based feature detection and matching methods to the stereo VO is time-consuming. Therefore, to improve efficiency, the following three techniques are utilized in our system. (1) For keyframes, we extract features on both left and right images and perform stereo matching to estimate the real scale. But for non-keyframes, we only process the left image. Besides, we use some lenient criteria to make the selected keyframes in our system very sparse, so the runtime and resource consumption of feature detection and matching in our system are close to that of a monocular system. (2) We convert the inference code of the CNN and GNN from Python to C++, and deploy them using ONNX and NVIDIA TensorRT, where the 16-bit floating-point arithmetic replaces the 32-bit floating-point arithmetic. (3) We design a multi-thread pipeline. A producer-consumer model is used to split the system into two main threads, i.e., the front-end thread and the backend thread. The front-end thread extracts and matches features while the backend thread performs the initial pose estimation, keyframe insertion, and local bundle adjustment.

\subsection{Feature Matching} \label{sec:feature_matching}

We use LightGlue \cite{lindenberger2023lightglue} to match keypoints.
For line features, most of the current VO and SLAM systems use the LBD algorithm \cite{zhang2013efficient} or tracking sample points to match them. 
However, the LBD algorithm extracts the descriptor from a local band region of the line, so it suffers from unstable line detection due to challenging illumination or viewpoint changes. Tracking sample points can match the line detected with different lengths in two frames, but current SLAM systems usually use optical flow to track the sample points, which have a bad performance when the light conditions change rapidly or violently. Some learning-based line feature descriptors \cite{Pautrat_Lin_2021_CVPR} are also proposed, however, they are rarely used in current SLAM systems due to the increased time complexity.

Therefore, to address both the effectiveness problem and efficiency problem, we design a fast and robust line-matching method for illumination-challenging conditions. First, we associate keypoints with line segments through their distances. Assume that $M$ keypoints and $N$ line segments are detected on the image, where the $i$-th keypoint is denoted as $\mathbf{p}_i=(x_i,y_i)$ and the $j$-th line segment is denoted as $\mathbf{l}_j=(A_j,B_j,C_j,x_{j,1},y_{j,1}, x_{j,2}, y_{j,2})$, where $(A_j,B_j,C_j)$ are line parameters of $\mathbf{l}_j$ and $(x_{j,1},y_{j,1}, x_{j,2}, y_{j,2})$ are the endpoints. We first compute the distance between $\mathbf{p}_i$ and $\mathbf{l}_j$ through:
\begin{equation}\label{eq:point_line_distance}
   d_{ij} = d \left(\mathbf{p}_i, \mathbf{l}_j \right) = \frac{\lvert A_j \cdot x_i + B_j \cdot y_i + C_j \rvert}{\sqrt{A_j^2 + B_j^2}} .
\end{equation}
If $d_{ij} < 3$ and the projection of $\mathbf{p}_i$ on the coordinate axis lies within the projections of line segment endpoints, i.e., $ \min({x}_{j,1}, {x}_{j,2}) \leq x_i \leq \max({x}_{j,1}, {x}_{j,2})$ or $ \min({y}_{j,1}, {y}_{j,2}) \leq {y}_i \leq \max({y}_{j,1}, {y}_{j,2})$, we will say $\mathbf{p}_i$ belongs to $\mathbf{l}_j$. Then the line segments on two images can be matched based on the point-matching result of these two images. For ${\mathbf{l}_{k, m}}$ on image $k$ and ${\mathbf{l}_{k+1, n}}$ on image $k+1$, we compute a score ${S}_{mn}$ to represent the confidence of that they are the same line:
\begin{equation}\label{eq:line_matching_score}
   {S}_{mn} = \frac{{N}_{pm}}{\min({{N}_{k, m}}, {{N}_{k+1, n}})},
\end{equation}
where ${N}_{pm}$ is the matching number between point features belonging to ${\mathbf{l}_{k, m}}$ and point features belonging to ${\mathbf{l}_{k+1, n}}$. ${{N}_{k, m}}$ and ${{N}_{k+1, n}}$ are the numbers of point features belonging to ${\mathbf{l}_{k, m}}$ and ${\mathbf{l}_{k+1, n}}$, respectively. Then if ${S}_{mn} > \delta_{S}$ and ${N}_{pm} > \delta_{N}$, where $\delta_{S}$ and $\delta_{N}$ are two preset thresholds, we will regard ${\mathbf{l}_{k, m}}$ and ${\mathbf{l}_{k+1, n}}$ as the same line.
This coupled feature matching method allows our line matching to share the robust performance of keypoint matching while being highly efficient due to that it does not need another line-matching network.

\subsection{3D Feature Processing}

In this part, we will introduce our 3D feature processing methods, including 3D feature representation, triangulation, \ie constructing 3D features from 2D features, and re-projection, \ie projecting 3D features to the image plane. 
For 3D point processing, a 3D point is denoted as $\mathbf{X} \in \mathbb{R}^3$. LightGlue \cite{lindenberger2023lightglue} is utilized to match keypoints between the left and right images. Successfully matched keypoints are triangulated using stereo disparity information, while unmatched keypoints are triangulated leveraging multi-frame observations. The projection of 3D points onto the image plane is modeled using either the pinhole camera model or the fisheye camera model, depending on the specific camera in use.
We skip the details of 3D point processing in our system as they are similar to other point-based VO and SLAM systems \cite{mur2017orb, qin2018vins}. 
On the contrary, compared with 3D points, 3D lines have more degrees of freedom, and they are easier to degenerate when being triangulated. Therefore, the 3D line processing will be illustrated in detail.

\subsubsection{3D Line Representation}\label{sec:3d_line_representation}
We use Pl\"{u}cker coordinates \cite{bartoli2005structure} to represent a 3D spatial line:
\begin{equation}\label{eq:line_plucker}
    \mathbf{L} = \left[\begin{array}{c} \mathbf{n} \\ \mathbf{v} \end{array}\right] \in \mathbb{R}^{6},
\end{equation}
where $\mathbf{v}$ is the direction vector of the line and $\mathbf{n}$ is the normal vector of the plane determined by the line and the origin. Pl\"{u}cker coordinates are used for 3D line triangulation, transformation, and projection. It is over-parameterized because it is a 6-dimensional vector, but a 3D line has only four degrees of freedom. In the graph optimization stage, the extra degrees of freedom will increase the computational cost and cause the numerical instability of the system \cite{zuo2017robust}. Therefore, we also use orthonormal representation \cite{bartoli2005structure} to represent a 3D line:
\begin{equation}\label{eq:line_orthonormal}
    \left( \mathbf{U}, \mathbf{W} \right) \in \text{SO(3)} \times \text{SO(2)}
\end{equation}

The relationship between Pl\"{u}cker coordinates and orthonormal representation is similar to $\text{SO(3)}$ and $\text{so(3)}$. Orthonormal representation can be obtained from Pl\"{u}cker coordinates by:
\begin{equation}\label{eq:line_plucker_orthonormal_conversion}
    \mathbf{L} = \left[ \mathbf{n} \mid \mathbf{v} \right] = \underbrace{ \left[\begin{array}{ccc} \frac{\mathbf{n}}{\lVert \mathbf{n} \lVert} & \frac{\mathbf{v}}{\lVert \mathbf{v} \lVert} & \frac{\mathbf{n} \times \mathbf{v}}{\lVert \mathbf{n} \times \mathbf{v} \lVert}  \end{array}\right]}_{\mathbf{U} \in \text{SO(3)}} \underbrace{\left[\begin{array}{cc} \lVert \mathbf{n} \lVert & \mathbf{0} \\ \mathbf{0} & \lVert \mathbf{v} \lVert \\ \mathbf{0} & \mathbf{0} \end{array}\right]}_{\Sigma_{3 \times 2}},
\end{equation}
where $\Sigma_{3 \times 2}$ is a diagonal matrix and its two non-zero entries defined up to scale can be represented by an $\text{SO(2)}$ matrix:
\begin{equation}\label{eq:line_orthonormal_W}
    \mathbf{W} = \frac{1}{\sqrt{ {\lVert \mathbf{n} \lVert}^2 + {\lVert \mathbf{v} \lVert}^2 }} \left[\begin{array}{cc} {\lVert \mathbf{n} \lVert} & -{\lVert \mathbf{v} \lVert} \\ {\lVert \mathbf{v} \lVert} & {\lVert \mathbf{n} \lVert} \end{array}\right] \in \text{SO(2)}.
\end{equation}
In practice, this conversion can be done simply and quickly with the QR decomposition.

\subsubsection{Triangulation}\label{sec:3d_line_triangulation}
Triangulation is to initialize a 3D line from two or more 2D line features. In our system, we use two methods to triangulate a 3D line. The first is similar to the line triangulation algorithm $B$ in \cite{yang2019visual}, where the pose of a 3D line can be computed from two planes. To achieve this, we select two line segments, $\mathbf{l}_1$ and $\mathbf{l}_2$, on two images, which are two observations of a 3D line. Note that the two images can come from the stereo pair of the same keyframe or two different keyframes. $\mathbf{l}_1$ and $\mathbf{l}_2$ can be back-projected and construct two 3D planes, $\mathbf{\pi}_1$ and $\mathbf{\pi}_2$. Then the 3D line can be regarded as the intersection of $\mathbf{\pi}_1$ and $\mathbf{\pi}_2$. 

However, triangulating a 3D line is more difficult than triangulating a 3D point, because it suffers more from degenerate motions\cite{yang2019visual}. Therefore, we also employ a second line triangulation method if the above method fails, where points are utilized to compute the 3D line. In \sref{sec:feature_matching}, we have associated point features with line features. So to initialize a 3D line, two triangulated points $\mathbf{X}_1$ and $\mathbf{X}_2$, which belong to this line and have the shortest distance from this line on the image plane are selected. Then the Pl\"{u}cker coordinates of this line can be obtained through:
\begin{equation}\label{eq:line_two_points_plucker_conversion}
    \mathbf{L} = \left[\begin{array}{c} \mathbf{n} \\ \mathbf{v} \end{array}\right] = \left[\begin{array}{c} \mathbf{X}_1 \times \mathbf{X}_2 \\ \frac{\mathbf{X}_1 - \mathbf{X}_2}{\lVert \mathbf{X}_1 - \mathbf{X}_2 \lVert} \end{array}\right].
\end{equation}
This method requires little extra computation because the selected 3D points have been triangulated in the point triangulating stage. It is very efficient and robust.

\subsubsection{Re-projection}\label{sec:3d_line_projection}
Re-projection is used to compute the re-projection errors.
We use Pl\"{u}cker coordinates to transform and re-project 3D lines. First, we convert the 3D line from the world frame to the camera frame:
\begin{equation}\label{eq:line_transformation}
    \mathbf{L}_c = \left[\begin{array}{c} \mathbf{n}_c \\ \mathbf{v}_c \end{array}\right] = \left[\begin{array}{cc} {\mathbf{R}}_{cw} & \left[{\mathbf{t}_{cw}} \right]_{\times}{\mathbf{R}_{cw}} \\ \mathbf{0} & {\mathbf{R}_{cw}} \end{array}\right]\left[\begin{array}{c} \mathbf{n}_w \\ \mathbf{v}_w \end{array}\right] = {\mathbf{H}_{cw}} {\mathbf{L}_w},
\end{equation}
where $\mathbf{L}_c$ and $\mathbf{L}_w$ are Pl\"{u}cker coordinates of 3D line in the camera frame and world frame, respectively. ${\mathbf{R}_{cw}} \in \text{SO(3)}$ is the rotation matrix from world frame to camera frame and ${\mathbf{t}_{cw}} \in \mathbb{R}^{3}$ is the translation vector. $\left[\cdot \right]_{\times}$ denotes the skew-symmetric matrix of a vector and ${\mathbf{H}_{cw}}$ is the transformation matrix of 3D lines from world frame to camera frame. 

Then the 3D line $\mathbf{L}_c$ can be projected to the image plane through a line projection matrix ${\mathbf{P}_c}$:
\begin{equation}\label{eq:line_re-projection}
    \mathbf{l} = \left[\begin{array}{c} A \\ B \\ C \end{array}\right] = {\mathbf{P}_c}{\mathbf{L}_c}_{[:3]}  = \left[\begin{array}{ccc} f_x & 0 & 0 \\ 0 & f_y & 0 \\ -f_yc_x & -f_xc_y & f_xf_y \end{array}\right] {\mathbf{n}_c},
\end{equation}
where $\mathbf{l} = \left[\begin{array}{ccc} A & B & C \end{array}\right]^\top$ is the re-projected 2D line on image plane. ${\mathbf{L}_c}_{[:3]}$ donates the first three rows of vector $\mathbf{L}_c$.

\subsection{Keyframe Selection} \label{sec:kf_selection}

Observing that the learning-based data association method used in our system is able to track two frames that have a large baseline, so different from the frame-by-frame tracking strategy used in other VO or SLAM systems, we only match the current frame with the last keyframe. We argue this strategy can reduce the accumulated tracking error. 

Therefore, the keyframe selection is essential for our system. On the one hand, as described in \sref{sec:vo_overview}, we want to make keyframes sparse to reduce the consumption of computational resources. On the other hand, the sparser the keyframes, the more likely tracking failure happens. To balance the efficiency and the tracking robustness, a frame will be selected as a keyframe if any of the following conditions is satisfied:
\begin{itemize}
    \item The tracked features are less than $\alpha_1 \cdot N_s$. 
    \item The average parallax of tracked features between the current frame and the last keyframe is larger than $\alpha_2 \cdot \sqrt{WH}$.
    \item The number of tracked features is less than $N_{kf}$.
\end{itemize}
In the above, $\alpha_1$, $\alpha_2$, and $N_{kf}$ are all preset thresholds.  $N_s$ is the number of detected features. $W$ and $H$ respectively represent the width and height of the input image.

\subsection{Local Graph Optimization} \label{sec:local_ba}

To improve the accuracy, we perform the local bundle adjustment when a new keyframe is inserted. $N_o$ latest neighboring keyframes are selected to construct a local graph, where map points, 3D lines, and keyframes are vertices and pose constraints are edges. We use point constraints and line constraints as well as IMU constraints if an IMU is accessible. Their related error terms are defined as follows. 

\subsubsection{Point Re-projection Error}\label{sec:mono_point_constraint}
If the frame $i$ can observe the 3D map point $\mathbf{X}_{p}$, then the re-projection error is defined as:
\begin{equation}\label{eq:mono_point_constraint}
    \mathbf{r}_{i, X_p} = {\tilde{\mathbf{x}}_{i, p}} - \pi \left( {\mathbf{R}_{cw}} {\mathbf{X}_{p}} + {\mathbf{t}_{cw}} \right),
\end{equation}
where ${\tilde{\mathbf{x}}_{i, p}}$ is the observation of ${\mathbf{X}_{p}}$ on frame $i$ and $\pi \left(\cdot \right)$ represents the camera projection. 

\subsubsection{Line Re-projection Error}\label{sec:mono_line_constraint}
If the frame $i$ can observe the 3D line $\mathbf{L}_{q}$, then the re-projection error is defined as:
\begin{subequations}\label{eq:mono_line_constraint}
    \begin{align}\label{eq:line_error}
        \mathbf{r}_{i, L_q} &= e_l \left({\tilde{\mathbf{l}}_{i,q}}, {\mathbf{P}_c} \left({{\mathbf{H}_{cw}} {\mathbf{L}_{q}}} \right)_{[:3]} \right) \in \mathbb{R}^{2} ,\\
        e_l \left({\tilde{\mathbf{l}}_{i, q}}, \mathbf{l}_{i, q} \right) &= \left[\begin{array}{cc} d \left(\tilde{\mathbf{p}}_{i, q1}, \mathbf{l}_{i,q} \right) & d \left(\tilde{\mathbf{p}}_{i, q2}, \mathbf{l}_{i,q} \right) \end{array}\right]^\top,
    \end{align}
\end{subequations}
where ${\tilde{\mathbf{l}}_{i, q}}$ is the observation of $\mathbf{L}_{q}$ on frame $i$, 
${\tilde{\mathbf{p}}_{i, q1}}$ and ${\tilde{\mathbf{p}}_{i, q2}}$ are the endpoints of ${\tilde{\mathbf{l}}_{i,q}}$, and
$d \left(\mathbf{p}, \mathbf{l} \right)$ is the distance between point $\mathbf{p}$ and line $\mathbf{l}$ which is computed through \eqref{eq:point_line_distance}. 

\subsubsection{IMU Residuals}\label{sec:imu_constraint}
We first follow \cite{forster2016manifold} to pre-integrate IMU measurements between the frame $i$ and the frame $j$:
\begin{subequations}\label{eq:imu_preinteration}
    \begin{align}
        \Delta{\mathbf{\tilde{R}}_{ij}} &= \prod\limits_{k=i}^{j-1} \text{Exp} \left(\left( \bm{\tilde{\omega}}_k - \mathbf{b}_k^g - \bm{\eta}_k^{gd} \right) \Delta t \right) ,\\
        \Delta{\mathbf{\tilde{v}}_{ij}} &= \sum\limits_{k=i}^{j-1} \Delta{\mathbf{\tilde{R}}_{ik}} \left( \mathbf{\tilde{a}}_k - \mathbf{b}_k^a - \bm{\eta}_k^{ad} \right) \Delta t, \\
        \Delta{\mathbf{\tilde{p}}_{ij}} &= \sum\limits_{k=i}^{j-1} \left( \Delta{\mathbf{\tilde{v}}_{ik}} \Delta t + \frac{1}{2} \Delta{\mathbf{\tilde{R}}_{ik}} \left( \mathbf{\tilde{a}}_k - \mathbf{b}_k^a - \bm{\eta}_k^{ad} \right) \Delta t^2 \right),
    \end{align}
\end{subequations}
where $\bm{\tilde{\omega}_k}$ and $\mathbf{\tilde{a}}_k$ are respectively the angular velocity and the acceleration. $\mathbf{b}_k^g$ and $\mathbf{b}_k^a$ are biases of the sensor and they are modeled as constants between two keyframes through $\mathbf{b}_k^g = \mathbf{b}_{k+1}^g$ and $\mathbf{b}_k^a = \mathbf{b}_{k+1}^a$. $\bm{\eta}_k^{gd}$ and $\bm{\eta}_k^{ad}$ are Gaussian noises. Then IMU residuals are defined as:
\begin{subequations}\label{eq:imu_residuals}
    \begin{align}
        \mathbf{r}_{\Delta{{R}_{ij}}} &= \text{Log} \left(\left( \Delta{\mathbf{\tilde{R}}_{ij}} \text{Exp} \left( \frac{\partial \Delta{\mathbf{R}_{ij}}}{\partial \mathbf{b}^g} \delta \mathbf{b}^g \right)  \right) ^\top \mathbf{R}_{i}^\top \mathbf{R}_{j} \right) ,\\ \notag
        \mathbf{r}_{\Delta{{v}_{ij}}} &= \mathbf{R}_{i}^\top \left(\mathbf{v}_{j} - \mathbf{v}_{i} - \mathbf{g} \Delta t_{ij}  \right) \\ &- \left( \Delta\tilde{\mathbf{v}}_{ij} + \frac{\partial \Delta{\mathbf{v}_{ij}}}{\partial \mathbf{b}^g} \delta \mathbf{b}^g + \frac{\partial \Delta{\mathbf{v}_{ij}}}{\partial \mathbf{b}^a} \delta \mathbf{b}^a \right), \\ \notag
        \mathbf{r}_{\Delta{{p}_{ij}}} &= \mathbf{R}_{i}^\top \left(\mathbf{p}_{j} - \mathbf{p}_{i} - \mathbf{v}_i \Delta t_{ij} - \frac{1}{2} \mathbf{g} \Delta t_{ij}^2  \right) \\ &- \left( \Delta\tilde{\mathbf{p}}_{ij} + \frac{\partial \Delta{\mathbf{p}_{ij}}}{\partial \mathbf{b}^g} \delta \mathbf{b}^g + \frac{\partial \Delta{\mathbf{p}_{ij}}}{\partial \mathbf{b}^a} \delta \mathbf{b}^a \right), \\
        \mathbf{r}_{b_{ij}} &= \left[\begin{array}{cc} \left(\mathbf{b}_j^g - \mathbf{b}_i^g \right)^\top & \left(\mathbf{b}_j^a - \mathbf{b}_i^a \right)^\top \end{array}\right]^\top
    \end{align}
\end{subequations}
where $\mathbf{g}$ is the gravity vector in world coordinates. In our system, we combine the initialization process in \cite{qin2018vins} and \cite{campos2021orb} to estimate $\mathbf{g}$ and initial values of biases.

The factor graph is optimized by the g2o toolbox \cite{kummerle2011g}. The cost function is defined as:
\begin{subequations}\label{eq:cost_function}
    \begin{align}
    \notag \mathbf{E} &= \sum {\lVert \mathbf{r}_{i, X_p} \lVert}^2_{\Sigma_P} + \sum {\lVert \mathbf{r}_{i, L_q} \lVert}^2_{\Sigma_L} + \sum {\lVert \mathbf{r}_{\Delta{{R}_{ij}}} \lVert}^2_{\Sigma_{\Delta{R}}} \\ &+ \sum {\lVert \mathbf{r}_{\Delta{{v}_{ij}}} \lVert}^2_{\Sigma_{\Delta{v}}} + \sum {\lVert \mathbf{r}_{\Delta{{p}_{ij}}} \lVert}^2_{\Sigma_{\Delta{p}}} + \sum {\lVert \mathbf{r}_{b_{ij}} \lVert}^2_{\Sigma_b}.
    \end{align}
\end{subequations}
We use the Levenberg-Marquardt optimizer to minimize the cost function. The point and line outliers are also rejected in the optimization if their corresponding residuals are too large.

\subsection{Initial Map} \label{sec:initial_map}

As described in \sref{sec:system_architecture}, our map is optimized offline. Therefore, keyframes, map points, and 3D lines will be saved to the disk for subsequent optimization when the visual odometry is finished. For each keyframe, we save its index, pose, keypoints, keypoint descriptors, line features, and junctions. The correspondences between 2D features and 3D features are also recorded. To make the map faster to save, load, and transfer across different devices, the above information is stored in binary form, which also makes the initial map much smaller than the raw data. For example, on the OIVIO dataset \cite{kasper2019benchmark}, our initial map size is only about 2\% of the raw data size.

\section{Map Optimization and Reuse}\label{sec:optimization_relocalization}
\subsection{Offline Map Optimization} \label{sec:optimization}
This part aims to process an initial map generated by our VO module and outputs the optimized map that can be used for drift-free relocalization.
Our offline map optimization module consists of the following several map-processing plugins.

\subsubsection{Loop Closure Detection} \label{sec:lc}
Similar to most current vSLAM systems, we use a coarse-to-fine pipeline to detect loop closures.
Our loop closure detection relies on DBoW2 \cite{galvez2012bags} to retrieve candidates and LightGlue \cite{lindenberger2023lightglue} to match features. We train a vocabulary for the keypoint detected by our PLNet on a database that contains 35k images. These images are selected from several large datasets \cite{torii2013visual, taira2018inloc, torii201524} that include both indoor and outdoor scenes. The vocabulary has 4 layers, with 10 nodes at each layer, so it contains 10,000 words. 

\underline{\emph{Coarse Candidate Selection:}}
This step aims to find three candidates most similar to a keyframe $\mathcal{K}_i$ from a set $\mathcal{S}_1 = \left \{\mathcal{K}_j \mid j < i \right \}$.
Note that we do not add keyframes with an index greater than $\mathcal{K}_i$ to the set because this may miss some loop pairs. We build a co-visibility graph for all keyframes where two are connected if they obverse at last one feature. All keyframes connected with $\mathcal{K}_i$ will be first removed from  $\mathcal{S}_1$. Then we compute a similarity score between $\mathcal{K}_i$ and each keyframe in $\mathcal{S}_1$ using DBoW2. Only keyframes with a score greater than $0.3 \cdot S_{max}$ will be kept in $\mathcal{S}_1$, where $S_{max}$ is the maximum computed score. After that, we group the remaining keyframes. If two keyframes can observe more than 10 features in common, they will be in the same group. For each group, we sum up the scores of the keyframes in this group and use it as the group score. Only the top 3 groups with the highest scores will be retained. Then we select one keyframe with the highest score within the group as the candidate from each group. These three candidates will be processed in the subsequent steps.

\underline{\emph{Fine Feature Matching:}}
For each selected candidate, we match its features with $\mathcal{K}_i$. Then the relative pose estimation with outlier rejection will be performed. 
The candidate will form a valid loop pair with $\mathcal{K}_i$ if the inliers exceed 50.

\subsubsection{Map Merging}
A 3D feature observed by both frames of a loop pair is usually mistakenly used as two features. Therefore, in this part, we aim to merge the duplicated point and line features observed by loop pairs. For keypoint features, we use the above feature-matching results between loop pairs. If two matched keypoints are associated with two different map points, they will be regarded as duplicated features and only one map point will be retained. The correspondence between 2D keypoints and 3D map points, as well as the connections in the co-visibility graph, will also be updated.

For line features, we first associate 3D lines and map points through the 2D-3D feature correspondence and 2D point-line association built in \sref{sec:feature_matching}. Then we detect 3D line pairs that associate with the same map points. If two 3D lines share more than 3 associated map points, they will be regarded as duplicated and only one 3D line will be retained.

\subsubsection{Global Bundle Adjustment}

We perform the global bundle adjustment (GBA) after merging duplicated features. The form of the residuals and loss functions is similar to that in \sref{sec:local_ba}; however, unlike \sref{sec:local_ba}, all keyframes and features are jointly optimized, and loop closure residuals are also incorporated into the optimization process. In the initial stage of optimization, the re-projection errors of merged features are relatively large due to the VO drift error, so we first iterate 50 times without outlier rejection to optimize the variables to a good rough position, and then iterate another 40 times with outlier rejection.

We find that when the map is large, the initial 50 iterations can not optimize the variables to a satisfactory position. To address this, we first perform pose graph optimization (PGO) before the global bundle adjustment if a map contains more than 80k map points. Only the keyframe poses will be adjusted in the PGO and the cost function is defined as follows:
\begin{equation}\label{eq:pgo}
    \mathbf{E}_{pgo} = \sum {\lVert \text{Log} \left( \Delta \tilde{\mathbf{T}}_{ij}^{-1} \mathbf{T}_{i}^{-1} \mathbf{T}_{j} \right) \lVert ^2_{\Sigma_{ij}}},
\end{equation}
where $\mathbf{T}_{i} \in \text{SE(3)}$ and $\mathbf{T}_{j} \in \text{SE(3)}$ are poses of $\mathcal{K}_i$ and $\mathcal{K}_j$, respectively. $\text{Log}(\cdot) = \text{log}(\cdot)^{{\vee}}: \text{SE(3)} \to \text{se(3)}$ is the Logarithm map proposed in \cite{wang2008nonparametric}. $\mathcal{K}_i$ and $\mathcal{K}_j$ should either be adjacent or form a loop pair. After the pose graph optimization, the positions of map points and 3D lines will also be adjusted along with the keyframes in which they are first observed. 

The systems with online loop detection usually perform the GBA after detecting a new loop, so they undergo repeated GBAs when a scene contains many loops. In contrast, our offline map optimization module only does the GBA after all loop closures are detected, allowing us to reduce the optimization iterations significantly compared with them.

\subsubsection{Scene-Dependent Vocabulary} \label{sec:junction_voc}
We train a junction vocabulary aiming to be used for relocalization. The vocabulary is built on the junctions of keyframes in the map so it is scene-dependent. Compared with the keypoint vocabulary trained in \sref{sec:lc}, the database used to train the junction vocabulary is generally much smaller, so we set the number of layers to 3, with 10 nodes in each layer. The junction vocabulary is tiny, \ie about 1 megabyte, as it only contains 1000 words. Its detailed usage will be introduced in \sref{sec:map_reuse}.

\subsubsection{Optimized Map}
we save the optimized map for subsequent map reuse. Compared with the initial map in \sref{sec:initial_map}, more information is saved such as the bag of words for each keyframe, the global co-visibility graph, and the scene-dependent junction vocabulary. In the meantime, the number of 3D features has decreased due to the fusion of duplicate map points and 3D lines. Therefore, the optimized map occupies a similar memory to the initial map.

\subsection{Map Reuse} \label{sec:map_reuse}
In this part, we present our illumination-robust relocalization using an existing optimized map. In most vSLAM systems, recognizing revisited places typically needs two steps: (1) retrieving $N_{kc}$ keyframe candidates and (2) performing feature matching and estimating relative pose. The second step is usually time-consuming, so selecting a proper $N_{kc}$ is very important. A larger $N_{kc}$ will reduce the system's efficiency while a smaller $N_{kc}$ may prevent the correct candidate from being recalled. For example, in the loop closing module of ORB-SLAM3 \cite{campos2021orb}, only the three most similar keyframes retrieved by DBoW2 \cite{GalvezTRO12} are used for better efficiency. It works well as two frames in a loop pair usually have a short time interval and thus the lighting conditions are relatively similar. But for challenging tasks, such as the day/night relocalization problem, retrieving so few candidates usually results in a low recall rate. However, retrieving more candidates needs to perform feature matching and pose estimation more times for each query frame, which makes it difficult to deploy for real-time applications.  

To address this problem, we propose an efficient multi-stage relocalization method to make the optimized map usable in different lighting conditions. Our insight is that if most of the false candidates can be quickly filtered out, then the efficiency can be improved while maintaining or even improving the relocalization recall rate. Therefore, we add another step to the two-step pipeline mentioned above. We next introduce the proposed multi-stage pipeline in detail.

\subsubsection{The First Step}
This step is to retrieve the similar keyframes in the map that are similar to the query frame. For each input monocular image, we detect keypoints, junctions, and line features using our PLNet. Then a pipeline similar to the ``coarse candidate selection" in \sref{sec:lc} will be executed, but with two differences. The first difference is that we do not filter out candidates using the co-visibility graph as the query frame is not in the graph. The second is that all candidates, not just three, will be retained for the next step.

\subsubsection{The Second Step} \label{sec:map_reuse_s2}
This step filters out most of the candidates selected in the first step using junctions and line features. For query frame $\mathcal{K}_q$ and each candidate $\mathcal{K}_b$, we first match their junctions by finding the same words through the junction vocabulary trained in \sref{sec:junction_voc}. We use $\left \{ \left(q_i, b_i \right) \mid q_i \in \mathcal{K}_q, b_i \in \mathcal{K}_b \right \}$ to denote the matching pairs. Then we construct two structure graphs, \ie $G_q^J$ and $G_b^J$, for $\mathcal{K}_q$ and $\mathcal{K}_b$, respectively. The vertices are matched junctions, \ie $V_q^J = \left \{ q_i \mid q_i \in \mathcal{K}_q \right \}$ and $V_b^J = \left \{ b_i \mid b_i \in \mathcal{K}_b \right \}$. The related adjacent matrices that describe the connection between vertices are defined as:
\begin{equation}\label{eq:adj}
    \mathbf{A}_q^J = \left[\begin{array}{ccc} q_{11} & \cdots & q_{1n} \\ \vdots & \ddots & \vdots \\ q_{n1} & \cdots & q_{nn} \end{array}\right],
    \mathbf{A}_b^J = \left[\begin{array}{ccc} b_{11} & \cdots & b_{1n} \\ \vdots & \ddots & \vdots \\ b_{n1} & \cdots & b_{nn} \end{array}\right],
\end{equation}
where $n$ is the number of junction-matching pairs. $q_{ij}$ is set to 1 if the junction $q_i$ and $q_j$ are two endpoints of the same line, otherwise, it is set to 0. The same goes for $b_{ij}$. Then the graph similarity of $G_q^J$ and $G_b^J$ can be computed through: 
\begin{equation}\label{eq:graph_sim}
    S_{qb}^G = \sum {1 - |q_{ij} - b_{ij}|}.
\end{equation}

We also compute a junction similarity score $S_{qb}^J$ using the junction vocabulary and the DBoW2 algorithm. Finally, the similarity score of $\mathcal{K}_q$ and $\mathcal{K}_b$ is given by combining the keypoint similarity, junction similarity, and structure graph similarity:
\begin{equation}\label{eq:overall_sim}
    S_{qb} = S_{qb}^K +  S_{qb}^J \cdot \left(1 + \frac{S_{qb}^G}{n} \right),
\end{equation}
where $S_{qb}^K$ is the keypoint similarity of $\mathcal{K}_q$ and $\mathcal{K}_b$ computed in the first step. We compute the similarity score with the query frame for each candidate, and only the top 3 candidates with the highest similarity scores will be retained for the next step. 

\textbf{Analysis:} We next analyze the second step. In the normal two-step pipeline that uses the DBoW method, only appearance information is used to retrieve candidates. The structural information, \ie the necessity of the consistent spatial distribution of features between the query frame and candidate, is ignored in the first step and only used in the second step. However, in the illumination-challenging scenes, the structural information is essential as it is invariant to lighting conditions. In our second step, a portion of the structural information is utilized to select candidates. First, our PLNet uses the wireframe-parsing method to detect structural lines, which are more stable in illumination-challenging environments. Second, the similarity computed in \eqref{eq:overall_sim} utilizes both the appearance information and the structural information. Therefore, our system can achieve good performance in illumination-challenging environments although using the efficient DBoW method.

The second step is also highly efficient. On the one hand, junctions are usually much less than keyponts. In normal scenes, our PLNet can detect more than 400 good keyponts but only about 50 junctions. On the other hand, the junction vocabulary is tiny and only contains 1,000 words. Therefore, matching junctions using DBoW2, constructing junction graphs, and computing similarity scores are all executed very efficiently. The experiment shows that the second step can be done within 0.7ms. More results will be presented in \sref{sec:experiments}.

\subsubsection{The Third Step}
The third step aims to estimate the pose of the query frame. We first use LightGlue to match features between the query frame and the retained candidates. The candidate with the most matching inliers will be selected as the best candidate. Then based on the matching results of the query frame and the best candidate, we can associate the query keypoints with map points. Finally, a PnP problem is solved with RANSAC to estimate the pose. The pose will be considered valid if the inliers exceed 20.

\section{Experiments} \label{sec:experiments}

In this section, we present the experiment results. 
The remainder of this section is organized as follows. In \sref{sec:line_e}, we evaluate the line detection performance of the proposed PLNet. In \sref{sec:mapping_acc}, we evaluate the mapping accuracy of our system by comparing it with other SOTA VO or SLAM systems. In \sref{sec:mapping_rob}, we test our system in three illumination-challenging scenarios: onboard illumination, dynamic illumination, and low illumination. The comparison of these three scenarios will show the excellent robustness of our system. In \sref{sec:map_reuse_e}, we assess the performance of the proposed map reuse module in addressing the day/night localization challenges, \ie mapping during the day and relocalization at night. In \sref{sec:ablation_study}, we present the ablation study. In \sref{sec:efficiency_analysis}, we evaluate the efficiency. 

We use two platforms in the experiments. Most evaluations are conducted on a personal computer with an Intel i9-13900 CPU and a NVIDIA GeForce RTX 4080 GPU. In the efficiency experiment in \sref{sec:efficiency_analysis}, we also deploy AirSLAM on an NVIDIA Jetson Orin to prove that our system can achieve good accuracy and efficiency on the embedded platform.

\subsection{Line Detection} \label{sec:line_e}
In this section, we evaluate the performance of our PLNet. As described in \sref{sec:fd_architecture}, we follow SuperPoint \cite{detone2018superpoint} to design and train our backbone and keypoint detection module, and we can even use the pre-trained model of SuperPoint, therefore, we do not evaluate the keypoint detection anymore. Instead, we assess the performance of the line detection module by comparing it with SOTA systems, as it is trained with a fixed backbone, which is different from other line detectors.    

\subsubsection{Datasets and Baseliens}
This experiment is conducted on the Wireframe dataset \cite{huang2018learning} and the YorkUrban dataset \cite{denis2008efficient}. The Wireframe dataset contains 5,000 training images and 462 test images that are all collected in man-made environments. We use them to train and test our PLNet. To validate the generalization ability, we also compare various methods on the YorkUrban dataset, which contains 102 test images. All the training and test images are resized to $512 \times 512$. 
We compare our method with AFM \cite{xue2019learning}, AFM++ \cite{xue2019learning_}, L-CNN \cite{zhou2019end}, LETR \cite{xu2021line}, F-Clip \cite{dai2022fully}, ELSD \cite{zhang2021elsd}, and HAWPv2 \cite{xue2023holistically}. 

\subsubsection{Evaluation Metrics}
We evaluate both the accuracy and efficiency of the line detection. For accuracy, the structural average precision (sAP) \cite{zhou2019end} is the most challenging metric of the wireframe parsing task. It is inspired by the mean average precision (mAP) commonly used in object detection. A detected line $\tilde{l} = \left(\tilde{\mathbf{p}}_1, \tilde{\mathbf{p}}_2\right)$ is a True Positive (TP) if and only if it satisfies the following:  
\begin{equation}\label{eq:sap}
    \min_{\left(\mathbf{p}_1, \mathbf{p}_2\right) \in \mathcal{L}} \| \mathbf{p}_1 - \tilde{\mathbf{p}}_1 \|^2 + \| \mathbf{p}_2 - \tilde{\mathbf{p}}_2 \|^2 \leq \vartheta,
\end{equation}
where $\mathcal{L}$ is the set of ground truth, and $\vartheta$ is a predefined threshold. We follow the previous methods to set $\vartheta$ to 5, 10, and 15, then the corresponding sAP scores are represented by $\text{sAP}^5$, $\text{sAP}^{10}$, and $\text{sAP}^{15}$, respectively. For efficiency, we use the frames per second (FPS) to evaluate various systems.

\begin{figure}[t]
    \vspace{0.3em}
    \centering
    \setlength{\abovecaptionskip}{-0.4cm}
    \includegraphics[width=0.99\linewidth]{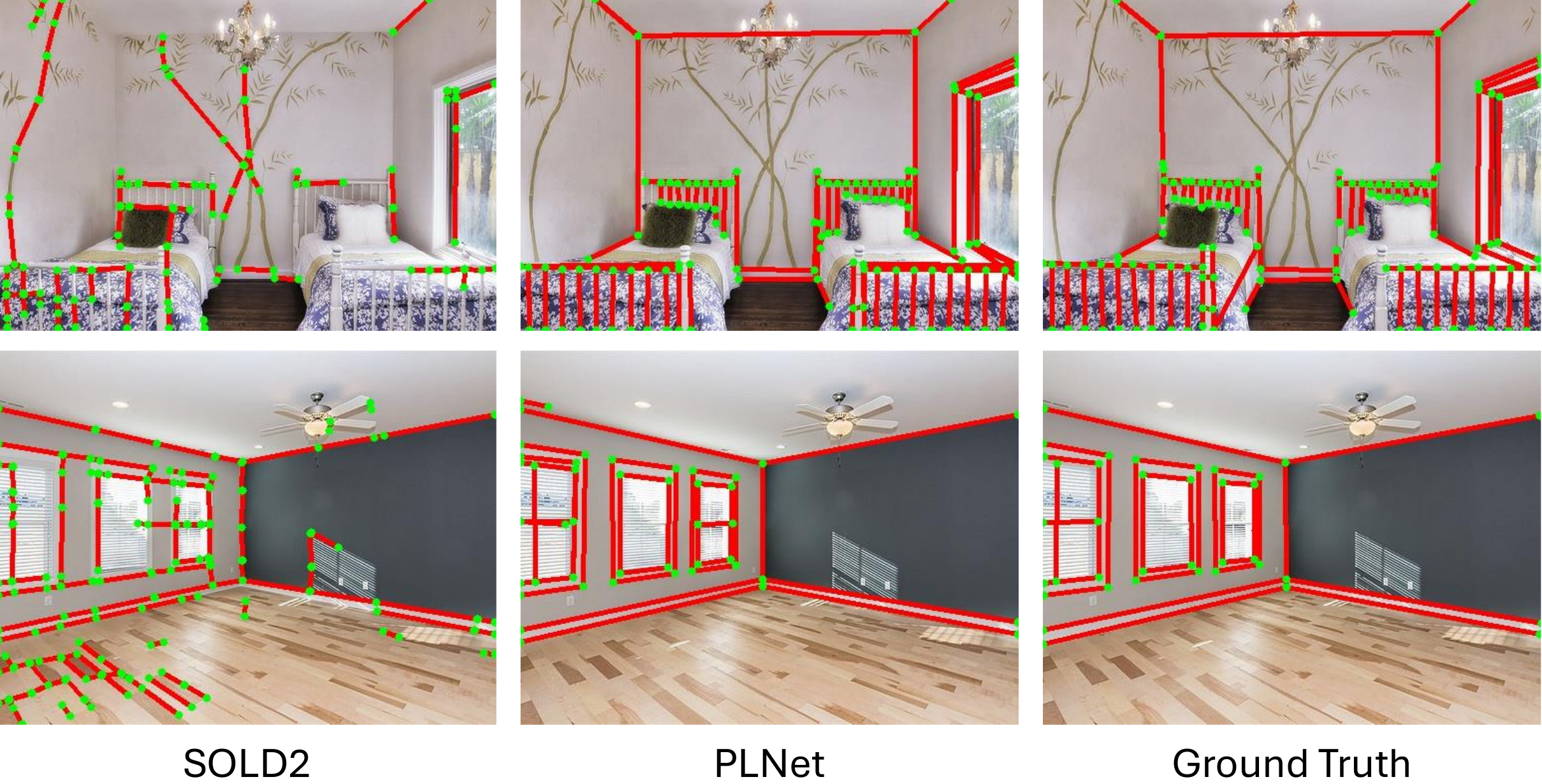}
    \caption{The line detection comparison between our PLNet (a wireframe parsing method) and SOLD2 (a non-wireframe-parsing method). The red lines are detected line features and the green points are endpoints of lines. Our PLNet aims to detect structural lines while SOLD2 detects more general lines with significant gradients, such as the patterns on the floor and walls.}
    \label{fig:sold2_plnet}
    \vspace{-0.2em}
\end{figure}

\begin{table}[t]
    \caption{The comparison of various wireframe parsing methods. The top two results are \textbf{highlighted} and \underline{underlined} in order.}
    \label{tab:line_detection}
    \centering
    \begin{threeparttable}
    \resizebox{\linewidth}{!}{
    \begin{tabular}{C{0.2\linewidth}|C{0.06\linewidth}C{0.06\linewidth}C{0.06\linewidth}|C{0.06\linewidth}C{0.06\linewidth}C{0.06\linewidth}|C{0.06\linewidth}}
        \toprule
        \multirow{2}{*}{Methods\tnote{1}} & \multicolumn{3}{c|}{Wireframe Dataset} & \multicolumn{3}{c|}{YorkUrban Dataset} & \multirow{2}{*}{FPS} \\
        \cline{2-7}
          & $\text{sAP}^5$ & $\text{sAP}^{10}$  & $\text{sAP}^{15}$ & $\text{sAP}^5$ & $\text{sAP}^{10}$  & $\text{sAP}^{15}$  &  \\
         \midrule
       \texttt{AFM} \cite{xue2019learning}     & 18.5 & 24.4 & 27.5 & 7.3 & 9.4 & 11.1 & 10.4\tnote{*}   \\
       \texttt{AFM++} \cite{xue2019learning_}   & 27.7 & 32.4 & 34.8 & 9.5 & 11.6 & 13.2 & 8.0\tnote{*}  \\
       \texttt{L-CNN} \cite{zhou2019end}   & 59.7 & 63.6 & 65.3 & 25.0 & 27.1 & 28.3 & 29.6  \\
       \texttt{LETR} \cite{xu2021line}    & 59.2 & 65.2 & 67.7 & 23.9 & 27.6 & 29.7 & 2.0  \\
       \texttt{F-Clip} \cite{dai2022fully}  & 64.3 & 68.3 & 69.1 & 28.6 & 31.0 & 32.4 & \underline{82.3}  \\
       \texttt{ELSD} \cite{zhang2021elsd}   & 64.3 & 68.9 & \underline{70.9} & 27.6 & 30.2 & 31.8 & 42.6\tnote{*}   \\
       \texttt{HAWPv2} \cite{xue2023holistically}    & \textbf{65.7} & \textbf{69.7} & \textbf{71.3} & \underline{28.9} & \underline{31.2} & \underline{32.6} & \textbf{85.2}  \\
        \midrule
       \textit{HAWPv2} \cite{xue2023holistically}    & 63.6 & 67.7 & 69.5 & 26.6 & 29.0 & 30.3 & 85.2  \\
       \textit{PLNet (Ours)}    & \underline{65.2} & \underline{69.2} & \underline{70.9} & \textbf{29.3} & \textbf{32.0} & \textbf{33.5} & 79.4\tnote{†}  \\
        \bottomrule
    \end{tabular}
    }
        \begin{tablenotes}
            \footnotesize
            \item[1] Methods represented using the \texttt{font} are evaluated with color inputs and methods represented using the \textit{font} are evaluated with grayscale inputs.
            \item[*] These numbers are cited from the original paper.
            \item[†] The FPS of our PLNet is the speed of detecting both keypoints and lines for the Python implementation.
        \end{tablenotes}
    \end{threeparttable}
    \vspace{-1.em}
\end{table}

\subsubsection{Results and Analysis}
We present the results in \tref{tab:line_detection}. The top-performing results are distinctly \textbf{highlighted} and \underline{underlined} in order.
It can be seen that our PLNet achieves the second-best performance on the Wireframe dataset and the best performance on the YorkUrban dataset. On the Wireframe dataset, HAWPv2, the best method, only outperforms our PLNet by 0.5, 0.5, and 0.4 points in $\text{sAP}^{5}$, $\text{sAP}^{10}$, and $\text{sAP}^{15}$, respectively. On the YorkUrban dataset, our method surpasses the second-best method by 0.4, 0.8, and 0.9 points on these three metrics, respectively. Overall, we can conclude that our PLNet achieves comparable accuracy with SOTA methods.

\begin{table*}[t]
    \caption{Translational error (RMSE) on the EuRoC dataset (unit: m), the best results are in \textbf{bold}.}
    \vspace{-5pt}
    \label{tab:euroc}
    \centering
    \begin{threeparttable}
    \resizebox{\linewidth}{!}{
    \begin{tabular}{C{0.01\linewidth}|C{0.165\linewidth}|C{0.008\linewidth}C{0.008\linewidth}C{0.008\linewidth}|C{0.008\linewidth}C{0.008\linewidth}|C{0.03\linewidth}C{0.03\linewidth}C{0.03\linewidth}C{0.03\linewidth}C{0.03\linewidth}C{0.03\linewidth}C{0.03\linewidth}C{0.03\linewidth}C{0.03\linewidth}C{0.03\linewidth}C{0.03\linewidth}C{0.03\linewidth}C{0.03\linewidth}C{0.03\linewidth}}
        \toprule
        \multirow{2}{*}{} & \multirow{2}{*}{} & \multicolumn{3}{c|}{Sensors} & \multicolumn{2}{c|}{Features} &  \multicolumn{11}{c}{Sequence} \\
         & & M\tnote{1} & S\tnote{1}  & I\tnote{1} & P\tnote{1} & L\tnote{1}  & MH01  & MH02 & MH03 & MH04 & MH05 & V101 & V102 & V103 & V201 & V202 & V203 & Avg\tnote{2}\\
         \midrule
        \multirow{6}{*}{\rotatebox{90}{Without Loop}}
       & DROID-SLAM \cite{teed2021droid} & \cc & \xx & \xx & \cc & \xx & 0.163 & 0.121 & 0.242 & 0.399 & 0.270 & 0.103 & 0.165 & 0.158 & 0.102 & 0.115 & 0.204  & 0.186 \\
       & VINS-Fusion \cite{qin2018vins} & \xx & \cc & \cc & \cc & \xx & 0.163 & 0.178 & 0.316 & 0.331 & 0.175 & 0.102 & 0.099 & 0.112 & 0.110 & 0.124 & 0.252  & 0.178 \\
       & Struct-VIO \cite{zou2019structvio} & \cc & \xx & \cc & \cc & \cc & 0.119 & 0.100 & 0.283 & 0.275 & 0.256 & 0.075 & 0.197 & 0.161 & 0.081 & 0.152 & 0.177  & 0.171 \\
       & PLF-VINS \cite{lee2021plf} & \xx & \cc & \cc & \cc & \cc & 0.143 & 0.178 & 0.221 & 0.240 & 0.260 & 0.069 & 0.099 & 0.166 & 0.083 & 0.125 & 0.183 & 0.161\\
       & Kimera-VIO \cite{rosinol2020kimera} & \xx & \cc & \cc & \cc & \xx & 0.110 &  0.100 & 0.160 & 0.240 & 0.350 & 0.050 & 0.080 & \textbf{0.070} & 0.080 & 0.100 & 0.210 & 0.141 \\
       & OKVIS \cite{leutenegger2015keyframe} & \xx & \cc & \cc & \cc & \xx & 0.197 & 0.108 & 0.122 & \textbf{0.138} & 0.272 & 0.040 & \textbf{0.067} & 0.120 & 0.055 & 0.150 & 0.240 & 0.137  \\
       \rowcolor{gray!20}
       & AirVIO (Ours) & \xx & \cc & \cc & \cc & \cc & \textbf{0.074} & \textbf{0.060} & \textbf{0.114} & 0.167 & \textbf{0.125} & \textbf{0.033} & 0.132 & 0.238 & \textbf{0.036} & \textbf{0.083} & \textbf{0.168} & \textbf{0.113}  \\  
         \midrule       
        \multirow{8}{*}{\rotatebox{90}{\makebox[-50pt][c]{With Loop}}}
       & iSLAM \cite{fu2024islam} & \xx & \cc & \cc & \xx & \xx & 0.302 & 0.460 & 0.363 & 0.936 & 0.478 & 0.355 & 0.391 & 0.301 & 0.452 & 0.416 & 1.133  & 0.508 \\
       & UV-SLAM \cite{lim2022uv} & \cc & \xx & \cc & \cc & \cc & 0.161 & 0.179 & 0.176 & 0.291 & 0.189 & 0.077 & 0.071 & 0.094 & 0.078 & 0.085 & 0.125 & 0.139 \\
       & Kimera \cite{rosinol2021kimera} & \xx & \cc & \cc & \cc & \xx & 0.090 &  0.110 & 0.120 & 0.160 & 0.180 & 0.050 & 0.060 & 0.130 & 0.050 & 0.070 & 0.230 & 0.114 \\
       & OpenVINS \cite{geneva2020openvins} & \xx & \cc & \cc & \cc & \xx & 0.072 & 0.143 & 0.086 & 0.173 & 0.247 & 0.055 & 0.060 & 0.059 & 0.054 & 0.047 & 0.141 & 0.103 \\
       & Structure-PLP-SLAM \cite{shu2023structure} & \xx & \cc & \xx & \cc & \cc & 0.046 & 0.056 & 0.048 & 0.071 & 0.071 & 0.091 & 0.066 & 0.065 & 0.061 & 0.061 & 0.166  & 0.073 \\
       & VINS-Fusion \cite{qin2018vins} & \xx & \cc & \cc & \cc & \xx & 0.052 & 0.040 & 0.052 & 0.124 & 0.088 & 0.046 & 0.053 & 0.108 & 0.040 & 0.081 & 0.098  & 0.071  \\
       & Maplab \cite{cramariuc2022maplab} & \cc & \xx & \cc & \cc & \xx & 0.041 & 0.026 & 0.045 & 0.110 & 0.067 & 0.039 & 0.045 & 0.080 & 0.053 & 0.084 & 0.196 & 0.071  \\
       & SP-Loop \cite{wang2023robust} & \xx & \cc & \cc & \cc & \xx & 0.070 & 0.044 & 0.068 & 0.100 & 0.090 & 0.042 & 0.034 & 0.082 & 0.038 & 0.054 & 0.100 & 0.066 \\
       & PL-SLAM \cite{gomez2019pl} & \xx & \cc & \cc & \cc & \cc & 0.042 & 0.052 & 0.040 & 0.064 & 0.070 & 0.042 & 0.046 & 0.069 & 0.061 & 0.057 & 0.126 & 0.061  \\
       & Basalt \cite{usenko2019visual} & \xx & \cc & \cc & \cc & \xx & 0.080 & 0.060 & 0.050 & 0.100 & 0.080 & 0.040 & 0.020 & 0.030 & 0.030 & 0.020 & 0.059 & 0.052 \\
       & DVI-SLAM \cite{peng2023dvi} & \xx & \cc & \cc & \cc & \xx & 0.042 & 0.046 & 0.081 & 0.072 & 0.069 & 0.059 & 0.034 & 0.028 & 0.040 & 0.039 & 0.055 & 0.051 \\
       & ORB-SLAM3 \cite{campos2021orb} & \xx & \cc & \cc & \cc & \xx & 0.036 & 0.033 & 0.035 & 0.051 & 0.082 & 0.038 & 0.014 & 0.024 & 0.032 & \textbf{0.014} & 0.024  & 0.035 \\
       & DROID-SLAM \cite{teed2021droid} & \xx & \cc & \xx & \cc & \xx & \textbf{0.015} & \textbf{0.013} & 0.035 & \textbf{0.048} & \textbf{0.040} & 0.037 & \textbf{0.011} & \textbf{0.020} & 0.018 & 0.015 & \textbf{0.017}  & \textbf{0.024} \\
       \rowcolor{gray!20}
       & AirSLAM (Ours) & \xx & \cc & \cc & \cc & \cc & 0.019 & \textbf{0.013} & \textbf{0.025} & 0.056 & 0.051 & \textbf{0.032} & 0.014 & 0.025 & \textbf{0.014} & 0.018 & 0.068 & 0.030  \\
          \bottomrule
    \end{tabular}
    }
        \begin{tablenotes}
            \footnotesize
            \item[1] M denotes the monocular camera, S denotes the stereo camera, I denotes the IMU, P denotes the keypoint feature, and L denotes the line feature.
            \item[2] The average error of the successful sequences.
        \end{tablenotes}
    \end{threeparttable}
    \vspace{-1.em}
\end{table*}

\paragraph*{Generalizability Analysis}
We can also conclude that the generalizability of our PLNet is better than other methods. This conclusion is based on two comparative results between our method and HAWPv2, which is the current best wireframe parsing method. First, on the Wireframe dataset, which also serves as the training dataset, HAWPv2 outperforms our PLNet. However, on the YorkUrban dataset, it is surpassed by our method. Second, the previous methods are all evaluated with color inputs in their original paper. Considering that grayscale images are also widely used in vSLAM systems, we train our PLNet with grayscale inputs. We also retrain HAWPv2 and evaluate it using grayscale images for comparison.
The result shows that our PLNet significantly outperforms HAWPv2 on both datasets when the inputs are grayscale images. We think the better generalizability comes from our backbone. Other methods are trained on only 5,000 images of the Wireframe dataset, while our backbone is trained on a large diverse dataset, which gives it a stronger feature extraction capability.

\paragraph*{Efficiency Analysis}
It is worth noting that the FPS of our method in \tref{tab:line_detection} is the speed of detecting both keypoints and lines, while other methods can only output lines. Nevertheless, our PLNet remains one of the fastest methods due to the design of the shared backbone. PLNet processes each image only $0.86 \milli\second$ slower than the fastest algorithm, \ie HAWPv2.

Note that the selected baselines are all wireframe parsing methods. The non-wireframe-parsing line detection methods, such as SOLD2 \cite{Pautrat_Lin_2021_CVPR} and DeepLSD \cite{pautrat2023deeplsd}, are not added to the comparison as it is unfair to do so. As shown in \fref{fig:sold2_plnet}, the wireframe parsing techniques aim to detect structural lines. They are usually evaluated using the sAP and compared with the ground truth. The non-wireframe-parsing methods can detect more general lines with significant gradients, however, they often detect a long line segment as multiple short line segments, which results in their poor sAP performance.

\begin{figure}[t]
    \vspace{0.1em}
    \centering
    \setlength{\abovecaptionskip}{-0.01cm}
    \includegraphics[width=0.92\linewidth]{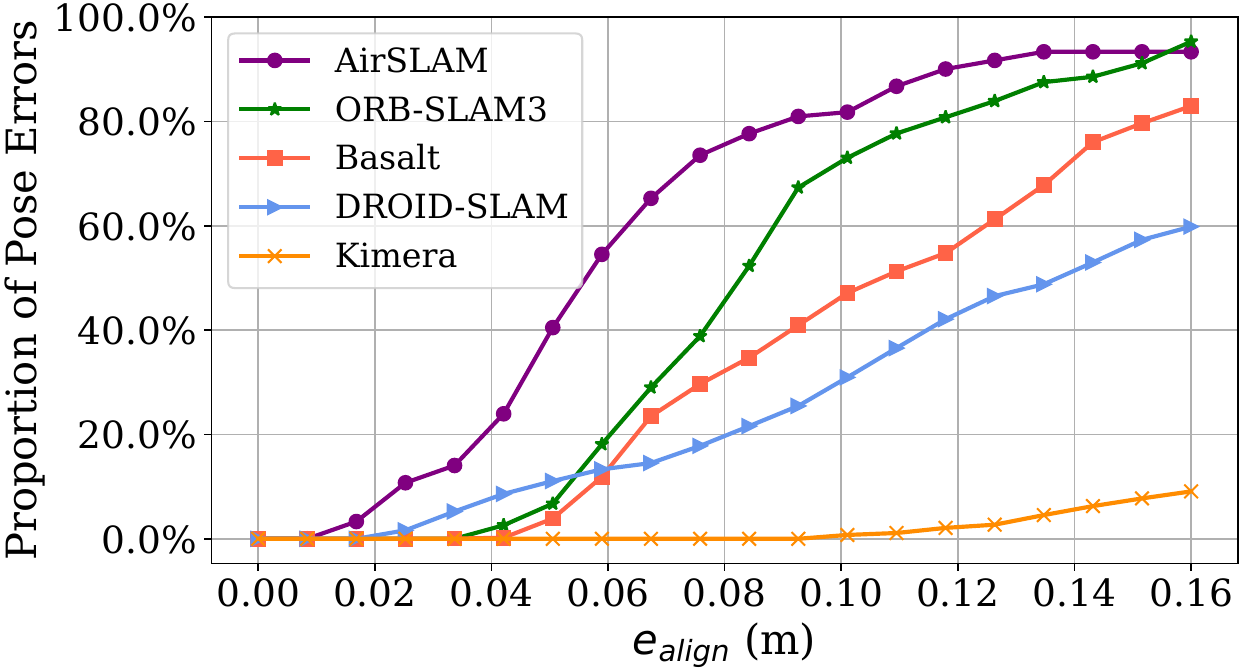}
     \caption{Comparison based on the OIVIO dataset. The vertical axis is the proportion of pose errors that are less than the given alignment error threshold on the horizontal axis. Our AirSLAM achieves the most accurate result.}
    \label{fig:oivio-rmse-curve}
    \vspace{-0.5em}
\end{figure}

\begin{table}[t]
    \vspace{0.0em}
    \setlength\tabcolsep{3.5pt}
    \caption{RMSE (m) on the OIVIO dataset, the best results are in \textbf{bold}. \textcolor{red}{F} represents tracking failure or large dirft error.}
    \label{tab:oivio_rmse}
    \centering  
    \begin{tabular}{C{0.22\linewidth}C{0.1\linewidth}C{0.09\linewidth}C{0.09\linewidth}C{0.11\linewidth}C{0.09\linewidth}C{0.09\linewidth}}
        \toprule
        \multirow{2}{*}{Sequence} & \multirow{2}{*}{Kimera} & PL- & \multirow{2}{*}{Basalt} & DROID- & ORB-  & \multirow{2}{*}{Ours} \\ 
        & & SLAM & & SLAM & SLAM3 \\
        \midrule
        MN\_015\_GV\_01 & 0.169              & 1.238 & 0.216 & 0.286  & 0.066  & \textbf{0.054}  \\
        MN\_015\_GV\_02 & 2.408              & 0.853 & 0.153 & 0.081  & 0.069  & \textbf{0.052} \\
        MN\_050\_GV\_01 & \textcolor{red}{F} & 1.143 & 0.186 & 0.173  & 0.063  & \textbf{0.062} \\
        MN\_050\_GV\_02 & \textcolor{red}{F} & 0.921 & 0.103 & 0.080 & 0.053  & \textbf{0.048} \\
        MN\_100\_GV\_01 & \textcolor{red}{F} & 0.831 & 0.197 & 0.184  & \textbf{0.051}  & 0.064 \\
        MN\_100\_GV\_02 & 2.238              & 0.609 & 0.092 & 0.090  & 0.063  & \textbf{0.042}  \\
        TN\_015\_GV\_01 & 0.300              & 1.579 & 0.148 & 0.188  & \textbf{0.053}  & 0.057  \\
        TN\_050\_GV\_01 & 0.280              & 1.736 & 0.521 & 0.313  & 0.082  & \textbf{0.065}  \\
        TN\_100\_GV\_01 & 0.264              & 1.312 & 0.116 & 0.179  & 0.086  & \textbf{0.078} \\
        \midrule
        Average       & - & 1.358 & 0.192 & 0.175 & 0.065 & \textbf{0.058} \\
        \bottomrule
    \end{tabular}
    \vspace{-1.em}
\end{table}

\subsection{Mapping Accuracy} \label{sec:mapping_acc}
In this section, we evaluate the mapping accuracy of our system under well-illuminated conditions. The EuRoC dataset \cite{burri2016euroc} is one of the most widely used datasets for vSLAM, so we use it for the accuracy evaluation.  
We compare our method only with systems capable of estimating the real scale, so the selected baselines are either visual-inertial systems, stereo systems, or those incorporating both. 
We incorporate traditional methods, learning-based systems, and hybrid systems into the comparison.
We use AirVIO to represent our system without loop detection.
The root mean square error (RMSE) is used as the metric and computed by the evo \cite{grupp2017evo}.

\begin{figure*}[t]
    \vspace{0.5em}
    \centering
    \includegraphics[width=0.98\linewidth]{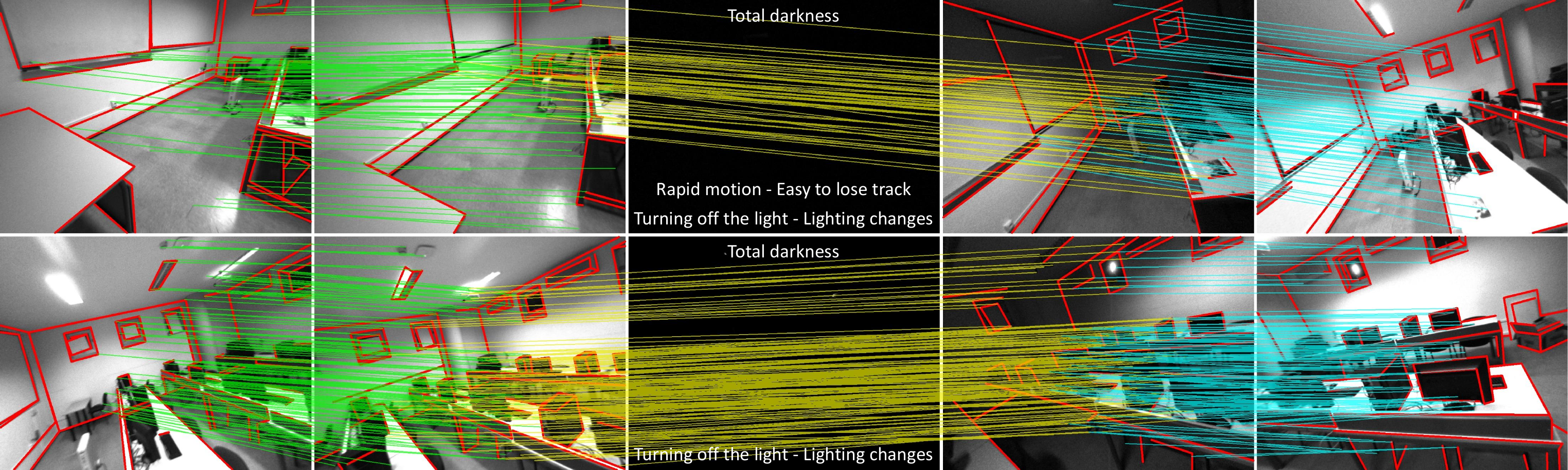}
    \caption{Our feature detection and matching on a challenging sequence in UMA-VI dataset. The red lines represent detected line features and the colored lines across images indicate feature association. The image may suddenly go dark due to turning off the lights, which is very difficult for vSLAM systems.}
    \label{fig:uma-seq}
    \vspace{-1.5em}
\end{figure*}

The comparison results are presented in \tref{tab:euroc}. We evaluate the systems with and without loop detection on 11 sequences. For the comparison without loop detection, our method outperforms other VIO methods: we achieve the best results on 8 out of 11 sequences. The average translational error of AirVIO is 20\% lower than the second-best system, \ie Kimera-VIO. For the comparison with loop detection, our system achieves comparable performance with ORB-SLAM3 and DROID-SLAM, which are SOTA traditional and learning-based visual SLAM systems, respectively.
Another conclusion that can be drawn from \tref{tab:euroc} is that loop detection significantly improves the accuracy of our system. The average error of our system decreases by 74\% after the loop detection.

\subsection{Mapping Robustness} \label{sec:mapping_rob}
Although many vSLAM systems have achieved impressive accuracy as shown in the previous \sref{sec:mapping_acc}, complex lighting conditions usually render them ineffective when deployed in real applications. Therefore, in this section, we evaluate the robustness of various vSLAM systems to lighting conditions. 
We select several representative SOTA systems as baselines. They are ORB-SLAM3 \cite{campos2021orb}, an accurate feature-based system, DROID-SLAM \cite{teed2021droid}, a learning-based hybrid system, Basalt \cite{usenko2019visual}, a system that achieves illumination-robust optical flow tracking with the LSSD algorithm, Kimera \cite{rosinol2021kimera}, a direct visual-inertial SLAM system, and OKIVS \cite{leutenegger2015keyframe}, a system proven to be illumination-robust in our previous work \cite{xu2023airvo}.
We test these methods and our system in three scenarios: onboard illumination, dynamic illumination, and low-lighting environments. We first present the evaluation results in \sref{sec:onboard_illumination}, \sref{sec:dynamic_illumination}, and \sref{sec:low_illumination}, respectively, and then give an overall analysis in \sref{sec:mapping_robostness_analysis}.

\begin{table}[t]
    \vspace{0.em}
    \setlength\tabcolsep{3.5pt}
    \caption{RMSE (m) on the UMA-VI dataset, the best results are in \textbf{bold}. \textcolor{red}{F} represents tracking failure or large dirft error.}
    \label{tab:uma_rmse}
    \centering  
    \begin{tabular}{C{0.25\linewidth}C{0.109\linewidth}C{0.09\linewidth}C{0.09\linewidth}C{0.08\linewidth}C{0.12\linewidth}C{0.08\linewidth}}
        \toprule
        \multirow{2}{*}{Sequence} & PL- & ORB- & \multirow{2}{*}{Basalt} & \multirow{2}{*}{OKVIS} & DROID-  & \multirow{2}{*}{Ours} \\ 
        & SLAM  &  SLAM3 & & & SLAM \\
        \midrule
        conference-csc1    & 2.697              & \textcolor{red}{F} & 1.270 & 1.118 & 0.711 & \textbf{0.490}  \\
        conference-csc2    & 1.596              & \textcolor{red}{F} & 0.682  & 0.470 & 0.135 & \textbf{0.091} \\
        conference-csc3    & \textcolor{red}{F} & 0.426              & 0.469 & \textbf{0.088} & 0.724 & \textbf{0.088} \\
        lab-module-csc-rev & \textcolor{red}{F} & \textbf{0.063}     & 0.486 & 0.861 & 0.364 & 0.504 \\
        lab-module-csc     & \textcolor{red}{F} & \textcolor{red}{F} & 0.403 & 0.579 & \textbf{0.319} & 0.979 \\
        long-walk-eng      & \textcolor{red}{F} & \textcolor{red}{F} & 5.046 & 3.005 & \textcolor{red}{F} & \textbf{1.801}  \\
        third-floor-csc1   & 4.478              & 0.863              & 0.420 & 0.287 & \textbf{0.048} & 0.070  \\
        third-floor-csc2   & 6.068              & 0.149              & 0.590 & 0.271 & 0.890 & \textbf{0.127}  \\
        two-floors-csc1    & \textcolor{red}{F} & \textcolor{red}{F} & 0.760 & 0.154 & 0.341 & \textbf{0.066} \\
        two-floors-csc2    & \textcolor{red}{F} & \textcolor{red}{F} & 1.211 & 0.679 & 0.299 & \textbf{0.190} \\
        \bottomrule
    \end{tabular}
    \vspace{0.3em}
\end{table}

\begin{figure}[t]
    \vspace{0.em}
    \centering
    \setlength{\abovecaptionskip}{-0.05cm}
    \includegraphics[width=0.96\linewidth]{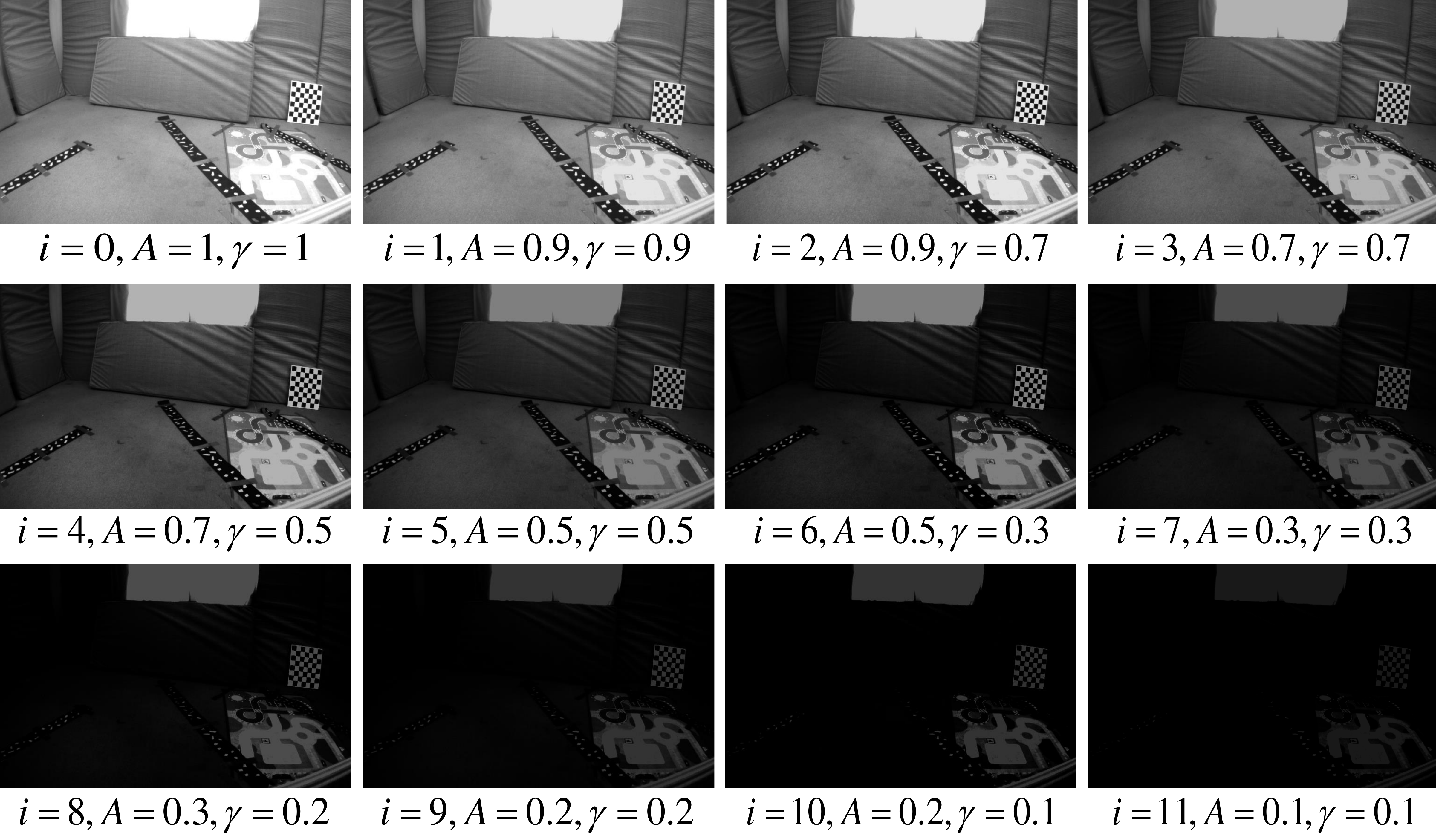}
    \caption{We use the gamma nonlinearity to generate image sequences with low illumination. $i$ is the brightness level.$A$ and $\gamma$ are parameters to control the image brightness. The smaller A and B are, the darker the image.}
    \label{fig:dark-euroc}
    \vspace{-1.0em}
\end{figure}

\subsubsection{Onboard Illumination} \label{sec:onboard_illumination}
We utilize the OIVIO dataset \cite{kasper2019benchmark} to assess the performance of various systems with onboard illumination. The OIVIO dataset collects visual-inertial data in tunnels and mines. In each sequence, the scene is illuminated by an onboard light of approximately 1300, 4500, or 9000 lumens. We used all nine sequences with ground truth acquired by the Leica TCRP1203 R300. As no loop closure exists in the selected sequences, it is fair to compare the VO systems with the SLAM systems. The performance of translational error is presented in \tref{tab:oivio_rmse}. The most accurate results are in \textbf{bold}, and \textcolor{red}{F} represents that the tracking is lost for more than $10\second$ or the RMSE exceeds $10\meter$. It can be seen that our method achieves the most accurate results on 7 out of 9 sequences and the smallest average error. The onboard illumination has almost no impact on our AirSLAM and ORB-SLAM3, however, it reduces the accuracy of OKVIS, Basalt, and PL-SLAM. Kimera suffers a lot from such illumination conditions. It even experiences tracking failures and large drift errors on three sequences.

We show a comparison of our method with selected baselines on the OIVIO TN\_100\_GV\_01 sequence in \fref{fig:oivio-rmse-curve}. In this case, the robot goes through a mine with onboard illumination. The distance is about 150 meters and the average speed is about 0.84m/s. The plot shows the proportion of pose errors on the horizontal axis that are less than the given alignment error threshold on the horizontal axis. Our system achieves a more accurate result than other systems on this sequence.

\begin{figure}[t]
    \vspace{0.em}
    \centering
    \setlength{\abovecaptionskip}{0.01cm}
    \includegraphics[width=0.96\linewidth]{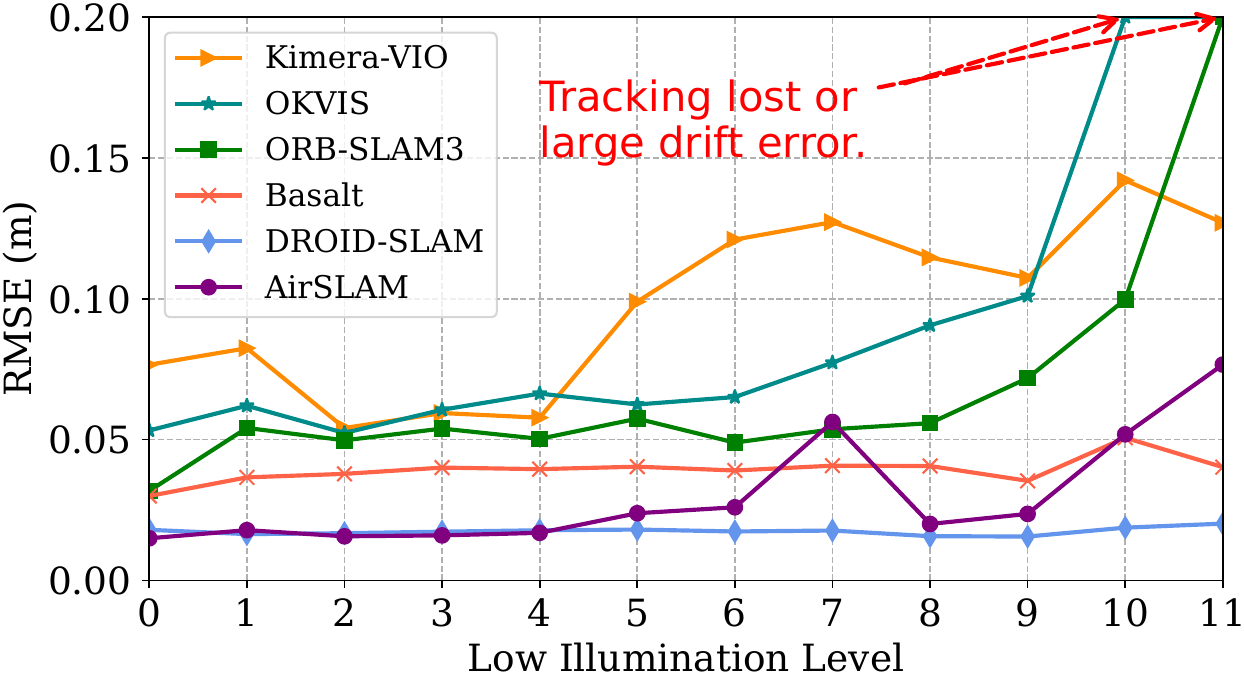}
    \caption{The comparison results on the Dark EuRoC dataset. The higher the level of low illumination on the x-axis, the darker the image. Basalt is the most stable system while our system is more accurate.}
    \label{fig:dark-euroc-error}
    \vspace{-1.0em}
\end{figure}

\subsubsection{Dynamic Illumination} \label{sec:dynamic_illumination}

The UMA-VI dataset is a visual-inertial dataset gathered in challenging scenarios with handheld custom sensors. We selected sequences with illumination changes to evaluate our system. As shown in \fref{fig:uma-seq}, it contains many subsequences where the image suddenly goes dark due to turning off the lights. In the subsequence of the top row in \fref{fig:uma-seq}, the total darkness lasted for 0.16\second, and low illumination lasted for 0.64\second. In the subsequence of the bottom row, the total darkness lasted for 0.32\second, and low illumination lasted for 0.4\second. It is more challenging than the OIVIO dataset for vSLAM systems. As the ground-truth poses are only available at the beginning and the end of each sequence, we disabled the loop closure part from all the evaluated methods.

The translational errors are presented in \tref{tab:uma_rmse}. The most accurate results are in \textbf{bold}, and \textcolor{red}{F} represents that the tracking is lost for more than $10\second$ or the RMSE exceeds $10\meter$. It can be seen that our AirSLAM outperforms other methods. Our system achieves the best results on 7 out of 10 sequences. The UMA-VI dataset is so challenging that PL-SLAM and ORB-SLAM3 fail on most sequences. Although OKVIS and Basalt, like our system, can complete all the sequences, their accuracy is significantly lower than ours. The average RMSEs of OKVIS and Basalt are around $1.134\meter$ and $0.724\meter$, respectively,  while ours is around $0.441\meter$, which means our average error is only 62.6\% of OKVIS and 38.9\% of Basalt.

\begin{figure}[t]
    \vspace{0.em}
    \centering
    \setlength{\abovecaptionskip}{-0.05cm}
    \includegraphics[width=0.96\linewidth]{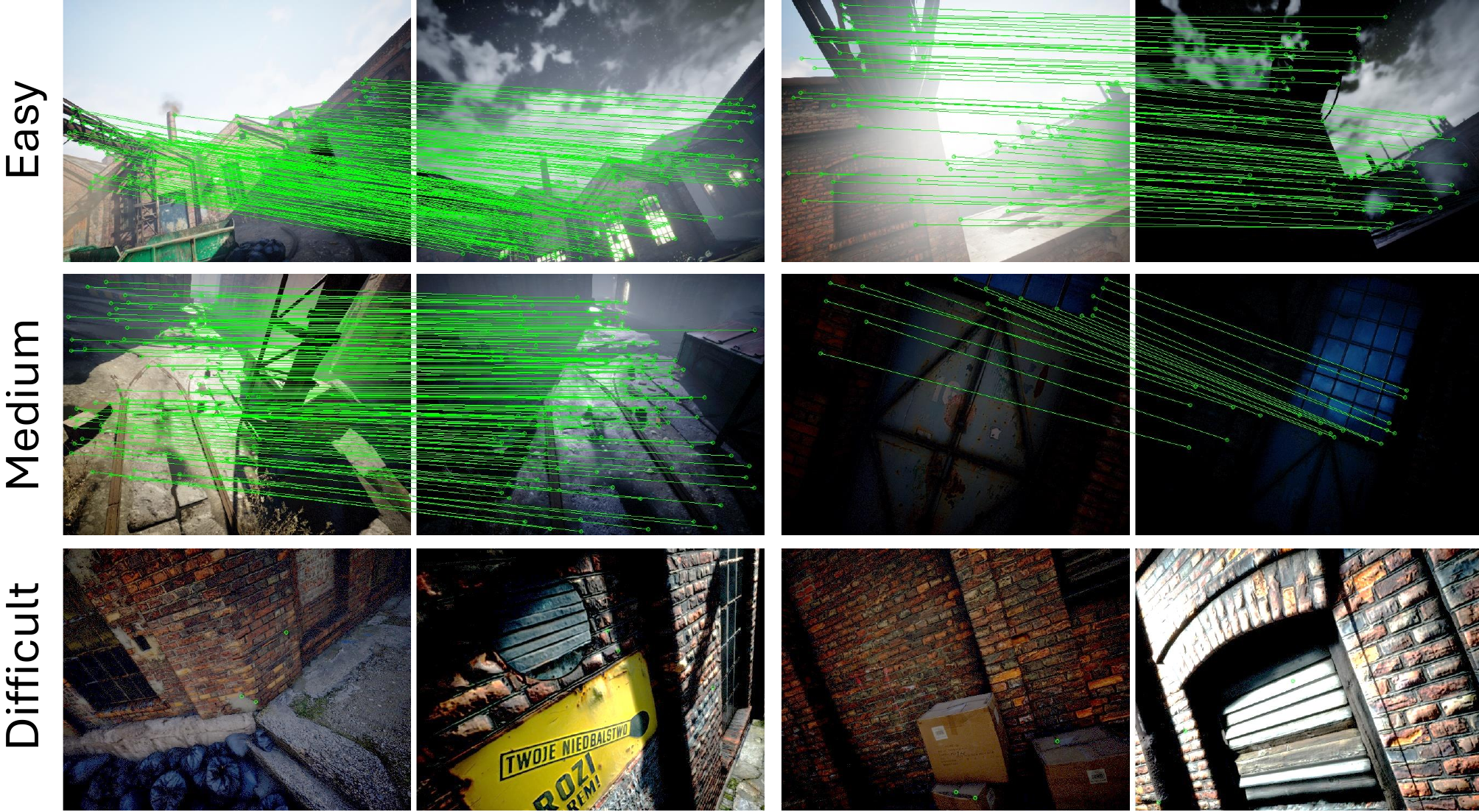}
    \caption{Some image pair samples for the mapping and relocalization in the TartanAir Day/Night Localization dataset. Due to the differences in capture viewpoints and scene depths, not all the image pairs have a valid overlap.}
    \label{fig:tartanair-sample}
    \vspace{-0.3em}
\end{figure}

\subsubsection{Low Illumination} \label{sec:low_illumination}
Inspired by \cite{park2017illumination}, we process a publicly available sequence by adjusting the brightness levels of its images. Then the processed sequences are used to evaluate the performance of various SLAM systems in low-illumination conditions. We select the ``V2\_01\_easy" of the EuRoC dataset as the base sequence. The image brightness is adjusted using the gamma nonlinearity: 
\begin{equation}\label{eq:gamma}
    V_{out} = AV_{in}^{\frac{1}{\gamma}},
\end{equation}
where $V_{in}$ and $V_{out}$ are normalized input and output pixel values, respectively. $A$ and $\gamma$ control the maximum brightness and contrast. We set 12 adjustment levels and use $L_i$ to denote the $i$th level. $L_0$ represents the original sequence, \ie $A_0=1$ and $\gamma_0=1$. When $i \in [1, 12]$, $A_i$ and $\gamma_i$ alternate in descending order to make the image progressively darker. \fref{fig:dark-euroc} shows the values of $A_i$ and $\gamma_i$, and the processed image in each level.
We name the processed dataset ``Dark EuRoC".

We present the comparison result in \fref{fig:dark-euroc-error}. As the errors of PL-SLAM are much greater than other methods, we do not show its result. Tracking failures and large drift errors, \ie the RMSE is more than 1\meter, are also marked. It can be seen that low illumination has varying degrees of impact on different systems. 
In this scenario, optical flow-based methods demonstrate greater stability compared to feature-based methods. Nonetheless, our approach achieves comparable performance to Basalt and DROID-SLAM, which utilize sparse and dense optical flow, respectively.
The RMSEs of ORB-SLAM3 and OKVIS increase as the brightness decreases. They even experience tracking failures or large drift errors on $L_{10}$ and $L_{11}$.

\begin{table*}[t]
    \caption{The relocalization comparison on the TartanAir Day/Night Localization dataset, the best results are in \textbf{bold}.}
    \vspace{-5pt}
    \label{tab:map_reuse}
    \centering
    \begin{threeparttable}
    \resizebox{\linewidth}{!}{
    \begin{tabular}{C{0.03\linewidth}|C{0.17\linewidth}|C{0.05\linewidth}|C{0.03\linewidth}C{0.03\linewidth}C{0.03\linewidth}C{0.03\linewidth}C{0.03\linewidth}C{0.03\linewidth}C{0.03\linewidth}C{0.03\linewidth}C{0.03\linewidth}C{0.03\linewidth}C{0.03\linewidth}C{0.03\linewidth}|C{0.03\linewidth}}
        \toprule
        \multirow{2}{*}{} & Global Feature\tnote{1} + & \multirow{2}{*}{FPS\tnote{2}} &  \multicolumn{11}{c}{Recall Rate of Sequences (\%)} \\
         & Matching\tnote{1} + Local Feature\tnote{1} &  & P000  & P001 & P002 & P003 & P004 & P005 & P006 & P007 & P008 & P009 & P010 & P011 & Avg\\
         \midrule
        \multirow{6}{*}{\rotatebox{90}{\makebox[-40pt][c]{Hloc\cite{sarlin2019coarse}}}}
       & NV + NN + SOSNet & 12.4 & 4.3 & 10.6 & 42.5 & 10.5 & 33.9 & 26.1 & 7.2 & 27.2 & 32.7 & 6.5 & 24.9  & 7.7 & 19.5 \\
       & NV + NN + D2-Net & 10.34 & 22.6 & 20.7 & 87.4 & 19.0 & 20.1 & 65.9 & 27.5 & 49.0 & 69.0 & 64.3 & 71.4  & 23.4 & 45.0 \\
       & NV + NN + R2D2 & 8.14 & 15.7 & 32.3 & 92.4 & 22.7 & 52.7 & 75.5 & 14.4 & 84.9 & 54.8 & 67.7 & 60.7 & 26.0 & 50.0 \\
       & NV + NN + SP & 30.2 & 69.2 &  32.8 & 88.7 & 17.2 & 53.5 & 72.6 & 31.1 & 83.8 & 72.6 & 69.6 & 89.5 & 34.0 & 59.6 \\
       & NV + AL + SOSNet & 6.1 & 29.0 & 30.3 & 55.6 & 19.6 & 49.3 & 42.4 & 12.7 & 45.8 & 38.9 & 25.2 & 47.0  & 11.3 & 34.0 \\
       & NV + LG + SIFT & 10.8 & 45.9 & 31.3 & 83.0 & 58.5 & 52.4 & 64.7 & 18.8 & 78.8 & 57.2 & 43.6 & 79.8 & 33.6 & 53.7 \\
       & NV + LG + DISK & 12.9 & 22.9 & 35.4 & 97.9 & 62.9 & 60.9 & 84.2 & 33.8 & 89.9 & 90.0 & 85.5 & 74.9 & 33.4 & 64.3  \\
       & NV + LG + SP & 20.6 & 94.7 & 35.4 & 98.4 & 70.9 & 63.5 & 85.8 & 33.8 & 95.9 & \textbf{97.1} & 91.4 & 99.2 & 49.8 & 76.3 \\
       & DIR + LG + SP & 19.1 & 87.2 & 28.8 & 99.5 & 69.8 & 64.2 & 85.2 & 32.8 & \textbf{96.7} & \textbf{97.1} & 88.1 & 96.0 & 50.4 & 74.7 \\
       & OpenIBL + LG + SP & 21.3 & 88.6 & 35.4 & 97.7 & 70.3 & 64.3 & \textbf{86.1} & 34.7 & 96.4 & 91.3 & \textbf{95.7} & \textbf{99.7} & 45.1 & 75.4 \\
       & EP + LG + SP & 22.3 & \textbf{99.7} & 36.4 & \textbf{100.0} & 68.3 & 64.5 & 85.8 & 34.4 & 96.5 & 94.0 & 85.1 & 96.9 & 45.1 & 75.6 \\
         \midrule       
       \multicolumn{2}{c|}{AirSLAM (Ours)} & \textbf{48.8} & 89.5 & \textbf{78.8} & 87.8 & \textbf{94.6} & \textbf{88.2} & 85.6 & \textbf{70.0} & 72.8 & 83.4 & 78.8 & 81.4 & \textbf{54.5} & \textbf{80.5}  \\
        \bottomrule
    \end{tabular}
    }
        \begin{tablenotes}
            \footnotesize
            \item[1] NV is NetVLAD, DIR is AP-GeM/DIR, EP is EigenPlace, NN is Nearest Neighbor Matching, AL is AdaLAM, LG is LightGlue, and SP is SuperPoint.
            \item[2] Running time of relocalization measured in Frame Per Second (FPS).
        \end{tablenotes}
    \end{threeparttable}
    \vspace{-1.em}
\end{table*}

\subsubsection{Result Analysis} \label{sec:mapping_robostness_analysis}
We think the above three lighting conditions affect a visual system in different ways. The OIVIO dataset collects sequences in dark environments with only onboard illumination, so the light source moves along with the robot, which results in two effects. On the one hand, the lighting is uneven in the environment. The direction the robot is facing and the area closer to the robot is brighter than other areas. The uneven image brightness may lead to the uneven distribution of features. On the other hand, when the robot moves, the lighting of the same area will change, resulting in different brightness in different frames. The assumption of brightness constancy in some systems will be affected in such conditions.  The UMA-VI dataset is collected under dynamic lighting conditions, where the dynamic lighting is caused by the sudden switching of lights or moving between indoor and outdoor environments. The image brightness variations in the UMA-VI dataset are much more intense than those in the OIVIO dataset, which may even make the extracted feature descriptor inconsistent in consecutive frames. In low-illumination environments, both the brightness and contrast of captured images are very low, making the vSLAM system more difficult to detect enough good features and extract distinct descriptors. 

We summarize the above experiment results with the following conclusions. First, the systems that use descriptors for matching are more robust than the direct methods in illumination-dynamic environments. On the OIVIO dataset, our AirSLAM and ORB-SLAM3 outperform the other systems significantly. On the UMA-VI dataset, our method and OKVIS achieve the best and the second-best results, respectively. This is reasonable as the brightness constancy assumption constrains the direct methods. Despite Basalt uses LSSD to enhance its optical flow tracking, its accuracy still decreases significantly in these two scenarios. Second, the direct methods are more stable in the low illumination environments. This is because descriptor-based SLAM systems rely on enough high-quality features and descriptors, which are difficult to obtain on low brightness and contrast images. The direct methods use corners that are easier to detect, so the low illumination has less impact on them. Third, thanks to the robust feature detection and matching, the illumination robustness of our system is far better than that of other systems. AirSLAM achieves relatively high accuracy in these three illumination-challenging scenarios.

\subsection{Map Reuse} \label{sec:map_reuse_e}

\subsubsection{Dataset}
As mapping and relocalization in the same well-illuminated environment are no longer difficult for many current vSLAM systems, we only evaluate our map reuse module under illumination-challenging conditions, \ie the day/night localization task. We use the ``abandoned\_factory" and ``abandoned\_factory\_night" scenes in the TartanAir dataset \cite{wang2020tartanair} as they can provide consecutive stereo image sequences for the SLAM mapping and the corresponding accurate ground truth for the evaluation. The images in these two scenes are collected during the day and at night, respectively. We use the sequences in the ``abandoned\_factory" scene to build maps. Then, for each mapping image, the images with a relative distance of less than $3\meter$ and a relative angle of less than $15\degree$ from it in the ``abandoned\_factory\_night" scene are selected as query images. We call the generated mapping and relocalization sequences the ``TartanAir Day/Night Localization" dataset. \fref{fig:tartanair-sample} shows some sample pairs for mapping and relocalization. It is worth noting that due to the differences in capture viewpoints and scene depths, the query image selected based on the relative distance and angle may not always have valid overlapping with the mapping images.

\subsubsection{Baseline}
We have tried several traditional vSLAM systems, \eg ORB-SLAM3 \cite{campos2021orb}, and SOTA learning-based one-stage relocalization methods, \eg ACE \cite{brachmann2023accelerated} on the TartanAir Day/Night Localization dataset, and find they perform badly: their relocalization recall rates are below 1\%. Therefore, we only present the comparison results of our systems and some VPR methods. The Hloc toolbox \cite{sarlin2019coarse} uses the structure from motion (SFM) method to build maps and has integrated many image retrieval methods, local feature extractors, and matching methods for localization. We mainly compare our system with these methods. Specifically, the NetVLAD \cite{arandjelovic2016netvlad},  AP-GeM/DIR \cite{revaud2019learning}, OpenIBL \cite{ge2020self}, and EigenPlaces \cite{berton2023eigenplaces} are used to extract global features, the SuperPoint \cite{detone2018superpoint}, SIFT \cite{lowe2004distinctive}, D2-Net \cite{dusmanu2019d2}, SOSNet \cite{tian2019sosnet}, R2D2 \cite{revaud2019r2d2}, and DISK \cite{tyszkiewicz2020disk} are used to extract local features, and the LightGlue \cite{lindenberger2023lightglue}, AdaLAM \cite{cavalli2020adalam} and Nearest Neighbor Matching are used to match features. We combine these methods into various ``global feature detection + local feature matcher + local feature detection" pipelines for the mapping and relocalization. We do not add DXSLAM\cite{li2020dxslam} to the comparison as it uses NetVLAD and SuperPoint with the binary descriptor, which has been included in the above pipelines.

\begin{figure}[t]
    \vspace{0.2em}
    \centering
    \setlength{\abovecaptionskip}{-0.05cm}
    \includegraphics[width=0.96\linewidth]{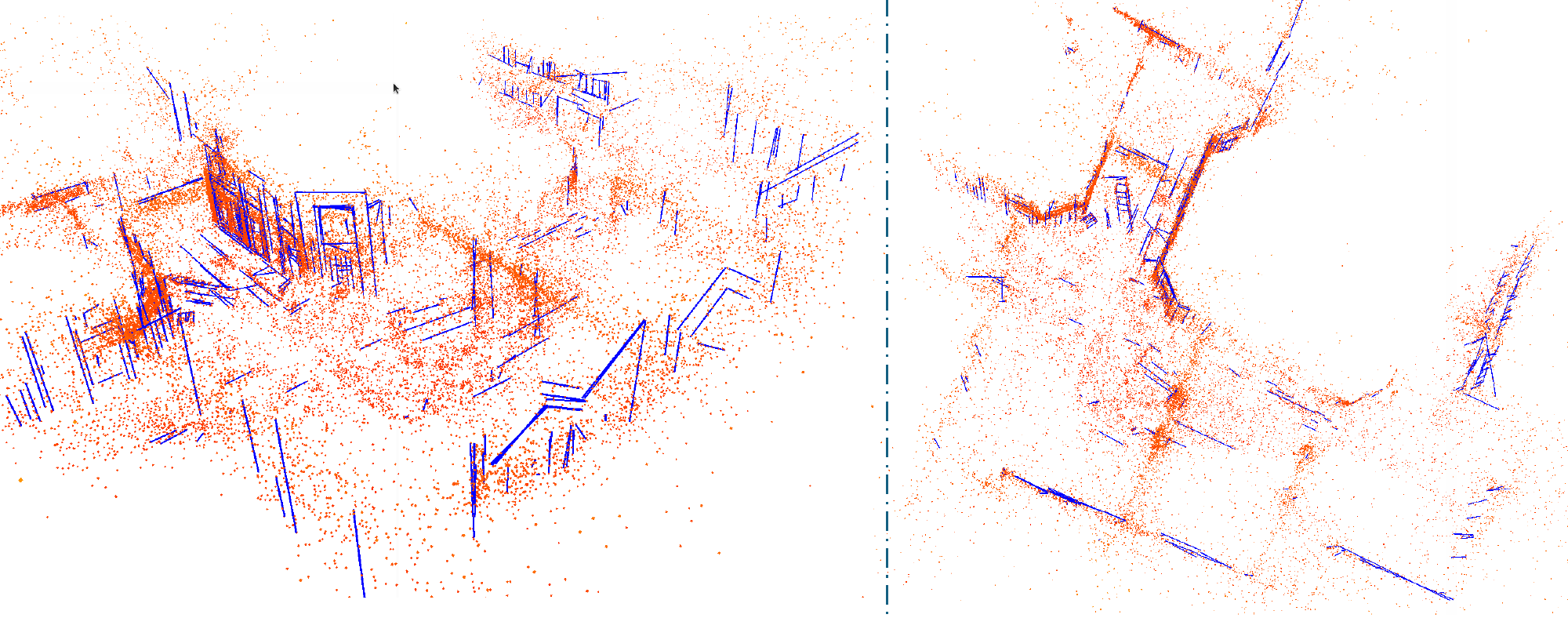}
    \caption{A point-line map of the P000 sequence built by our AirSLAM. The red points are mappoints and the blue lines are 3D lines.}
    \label{fig:tartanair_map}
    \vspace{-0.5em}
\end{figure}

\subsubsection{Results}
To achieve a fair comparison and balance the efficiency and effectiveness, we extract 400 local features and retrieve 3 candidates in the coarse localization stage for all methods. Unlike vSLAM systems that have the keyframe selection mechanism, the SFM mapping optimizes all input images, so it is very slow when mapping with original sequences. Therefore, to accelerate the SFM mapping while ensuring its mapping frames are more than our keyframes, we sample its mapping sequences by selecting one frame every four frames. We show a point-line map built by our AirSLAM in \fref{fig:tartanair_map}. The relocalization results are presented in \tref{tab:map_reuse}. We give the running time (FPS) and the relocalization recall rate of each method. We define a successful relocalization if the estimated pose of the query frame is within $2\meter$ and $15\degree$ of the ground truth. It can be seen that our AirSLAM outperforms other methods in terms of both efficiency and recall rate. Our system achieves the best results on 5 out of 11 sequences. AirSLAM has an average recall rate 4.2\% higher than the second-best algorithm and is about 2.4 times faster than it.

\begin{table}[t]
    \vspace{0.em}
    \setlength\tabcolsep{3.5pt}
    \caption{Ablation study of line features for mapping. We present the RMSE (m) of our system with and without line features.}
    \label{tab:ab_line}
    \centering  
    \begin{tabular}{C{0.2\linewidth}|C{0.13\linewidth}C{0.13\linewidth}C{0.13\linewidth}C{0.17\linewidth}}
        \toprule
        \multirow{2}{*}{}  & EuRoC & OIVIO & UMA-VI & Dark-EuRoC  \\
        \midrule
        w/o Lines    & 0.046 & 0.072 & 0.545 & 0.035 \\
        Ours  & \textbf{0.030} & \textbf{0.058} & \textbf{0.441} & \textbf{0.029}\\
        \bottomrule
    \end{tabular}
    \vspace{-0.5em}
\end{table}

\subsubsection{Analysis}
We find that our system is more stable than the VPR methods on the TartanAir Day/Night Localization dataset. On several sequences, \eg P000, P002, and P010, some VPR methods achieve remarkable results, with recall rates close to 100\%. However, on some other sequences, \eg P001 and P006, their recall rates are less than 40\%. In contrast, our system maintains a recall rate of 70\% to 90\% on most sequences.

To clarify this, we examined each sequence and roughly categorized the images into three types. As shown in \fref{fig:tartanair-sample}, the first type of image is captured with the camera relatively far from the features, so there is a significant overlap between each day/night image pair. Additionally, these images contain distinct buildings and landmarks. The second type of image pair also has a significant overlap. However, the common regions in these image pairs do not contain large buildings and landmarks. The third type of image is captured with the camera very close to the features. Although the camera distance between the day/night image pair is not large, they have almost no overlap, making their local feature matching impossible.

We find that the VPR methods perform very well on the first type of image but perform poorly on the second type of image. Therefore, their recall rates are very low on P001 and P006, which contain more of the second type of images. This may be because their global features are usually trained on datasets that have a lot of distinct buildings and landmarks, which makes them rely more on such semantic cues to retrieve similar images. By contrast, our system is based on the DBoW method, which only utilizes the low-level local features of images, so it achieves similar performance on the first and second types of images. This also proves the strong generalizability of our system. However, neither our system nor VPR methods can process the pairs with little overlap due to relying on local feature matching. Such image pairs are abundant in the P011.

\begin{table}[t]
    \caption{Relocalization ablation study. The recall rates (\%) of our system with and without the structure graph during map reuse.}
    \label{tab:ablation_map_reuse}
    \centering
    \begin{threeparttable}
    \begin{tabular}{C{0.12\linewidth}|C{0.08\linewidth}R{0.08\linewidth}|C{0.08\linewidth}C{0.08\linewidth}|C{0.08\linewidth}C{0.08\linewidth}}
        \toprule
        \multirow{2}{*}{Seq.} & \multicolumn{2}{c|}{$N_\mathcal{C}$=3} & \multicolumn{2}{c|}{$N_\mathcal{C}$=5}     & \multicolumn{2}{c}{$N_\mathcal{C}$=10}   \\ 
          & \text{w/o G.}   & Ours & \text{w/o G.} & Ours & \text{w/o G.}   & Ours       \\ 
        \midrule
        P000     & 77.9 & \textbf{89.5} & 84.3 & \textbf{92.5}     & 88.5  & \textbf{94.0} \\
        P001     & 69.2 & \textbf{78.8} & 79.3 & \textbf{83.8}     & 85.4  & \textbf{87.9} \\
        P002     & 75.4 & \textbf{87.8} & 80.9 & \textbf{87.8}     & 85.3  & \textbf{88.0}  \\ 
        P003     & 86.4 & \textbf{94.6} & 92.4 & \textbf{95.5}     & 93.9  & \textbf{95.9} \\ 
        P004     & 81.5 & \textbf{88.2} & 85.3 & \textbf{90.9}     & 89.0  & \textbf{92.1}  \\ 
        P005     & 78.4 & \textbf{85.6} & 82.9 & \textbf{88.8}     & 88.4  & \textbf{91.5}  \\ 
        P006     & 62.4 & \textbf{70.0} & 72.3 & \textbf{72.7}     & 78.6  & \textbf{77.2}  \\ 
        P007     & 60.6 & \textbf{72.8} & 63.7 & \textbf{73.7}     & 69.0  & \textbf{76.2} \\ 
        P008     & 77.2 & \textbf{83.4} & 80.0 & \textbf{84.4}     & 81.7  & \textbf{85.8}  \\ 
        P009     & 63.4 & \textbf{78.8} & 70.6 & \textbf{81.7}     & 76.9  & \textbf{84.8}  \\ 
        P010     & 70.7 & \textbf{81.4} & 75.4 & \textbf{84.0}     & 80.9  & \textbf{86.9}  \\ 
        P011     & 48.1 & \textbf{54.5} & 55.1 & \textbf{58.9}     & 62.0  & \textbf{64.0} \\
        \midrule
        Avg.     & 70.9 & \textbf{80.5} & 76.9 & \textbf{82.9}     & 81.6  & \textbf{85.4} \\
        \bottomrule     
    \end{tabular}
    \end{threeparttable}
    \vspace{-0.8em}
\end{table}

\subsection{Ablation Study} \label{sec:ablation_study}
\subsubsection{Line Feature In Mapping} \label{sec:ab_line}
In this section, we evaluate the impact of line features on mapping performance. To this end, we remove line features from our system and compute the RMSE on the EuRoC, OIVIO, UMA-VI, and Dark-EuRoC datasets. The results are summarized in \tref{tab:ab_line}, where ``\text{w/o Lines}" refers to our system without line features. As shown in the table, incorporating line features reduces the RMSE by 34.8\%, 19.4\%, 19.1\%, and 17.1\% on the four datasets, respectively. These results demonstrate that line features substantially enhance our system and improve the mapping accuracy. 
We believe this improvement comes from two aspects. On the one hand, a line feature is more likely to be observed across multiple frames compared to a point feature, thereby constraining more frames during pose optimization. This consistent constraint across multiple frames enhances the accuracy of pose estimation. On the other hand, we use the tracking results of a set of points to track a line feature, which makes the tracking of a single line feature more stable than that of a single point.

\subsubsection{Structure Graph In Relocalization} \label{sec:ab_sg}
We also verify the effectiveness of the proposed relocalization method. This experiment is conducted on the TartanAir Day/Night Localization dataset. We compare the systems with and without the second step proposed in \sref{sec:map_reuse_s2}. The results are presented in \tref{tab:ablation_map_reuse}, where ``\text{w/o G.}" denotes our system without the structure graph, and $N_\mathcal{C}$ denotes the candidate number for local feature matching. It shows that using junctions, line features, and structure graphs to filter out relocalization candidates significantly improves recall rates. AirSLAM outperforms \text{w/o G.} across all sequences, and when $N_\mathcal{C}$ is 3, 5, and 10, the average improvements are 9.6\%, 6.0\%, and 3.8\%, respectively, which demonstrates the effectiveness of the proposed method.

\subsection{Efficiency Analysis} \label{sec:efficiency_analysis}

Efficiency is essential for robotics applications, so we also evaluate the efficiency of the proposed system. We first compare the running time of our AirSLAM with several SOTA VO and SLAM systems on a computer with an Intel i9-13900 CPU and an NVIDIA RTX 4080 GPU. Then we deploy AirSLAM on an NVIDIA Jetson AGX Orin to verify the efficiency and performance of our system on the embedded platform.

\subsubsection{Odometry Efficiency}

\begin{figure}[t]
    \centering
    \subfloat[Odometry Efficiency.]
    {\includegraphics[width=0.48\linewidth]{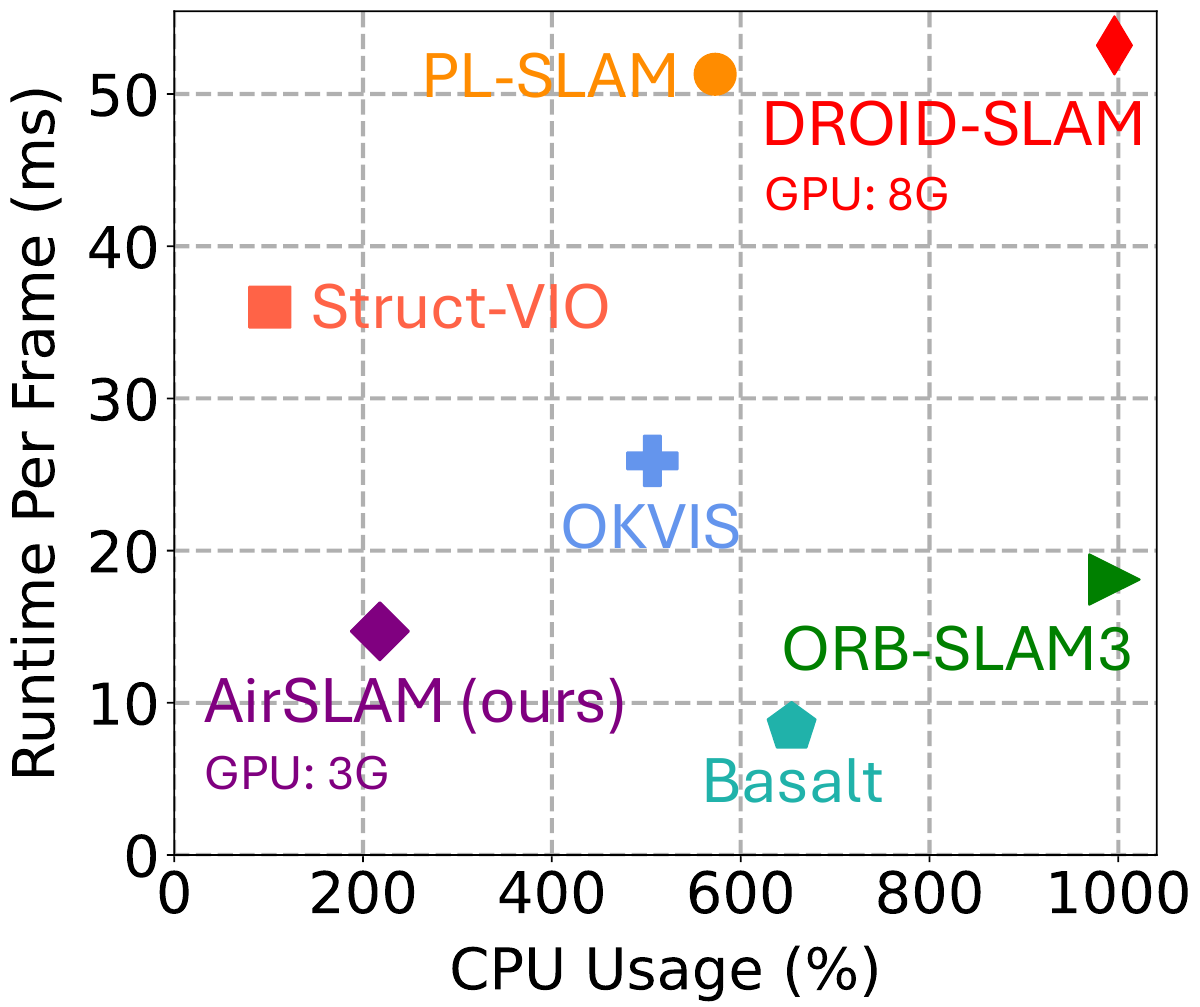}%
        \label{fig:vo_time}
    }
    \hfil
    \subfloat[Mapping Efficiency.]
    {\includegraphics[width=0.48\linewidth]{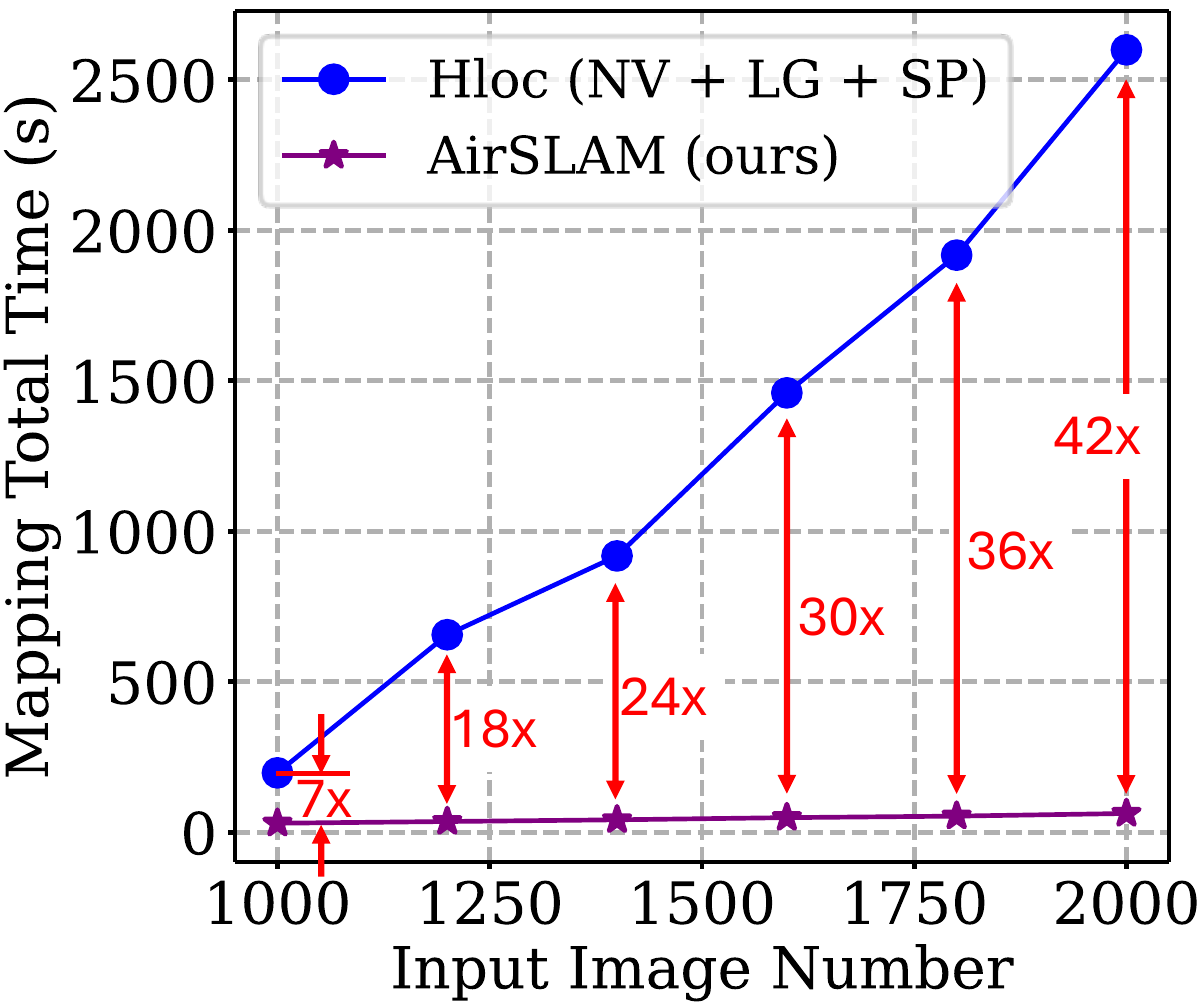}%
        \label{fig:mapping_time}
    }
    \caption{For odometry efficiency, we disable the loop detection from all the methods. The GPU usage of systems using GPU is also indicated. For mapping efficiency, we compute the total mapping time and compare it with Hloc.}
    \vspace{-0.7em}
\end{figure}

The VO/VIO efficiency experiment is conducted on the MH\_01\_easy sequence of the EuRoC dataset. We compare our AirSLAM with several SOTA systems. The loop detection and GBA are disabled from all the systems for a fair comparison. The metrics are the runtime per frame and the CPU usage. The results are presented in \fref{fig:vo_time}, where 100\% CPU usage means 1 CPU core is utilized. It should be additionally noted that DROID-SLAM actually uses 32 CPU cores, and its CPU usage in \fref{fig:vo_time} is only for a compact presentation.
Our system is very efficient, achieving a rate of 73 FPS. In addition, due to extracting and matching features using the GPU, our system requires relatively less CPU resources. 
We also test the GPU usage. It shows that DROID-SLAM requires about 8GB GPU memory, while our AirSLAM only requires around 3GB.

To investigate the contributions of the CPU and GPU to the efficiency of our system, we measured the \textbf{total runtime} of modules executed on the GPU and CPU using the MH\_01\_easy sequence and compared it with ORB-SLAM3. The loop detection and GBA are disabled from both systems.
The results are presented in \tref{tab:cpu_gpu}. It can be seen that our feature detection and matching are highly efficient due to the use of the GPU resources and our system design, which processes only the left image in non-keyframes. Note that we perform the initial pose estimation in the back-end, whereas ORB-SLAM3 includes this process in the front-end. This leads to the differences in the runtime of the ``Others" and ``Back-end" components between the two systems.

\begin{table}[t]
    \vspace{0.0em}
    \caption{Evaluation of CPU and GPU contributions to our system.}
    \label{tab:cpu_gpu}
    \centering  
    \begin{threeparttable}
    \begin{tabular}{C{0.2\linewidth}|C{0.14\linewidth}C{0.14\linewidth}|C{0.14\linewidth}}
        \toprule
        \multirow{2}{*}{} & \multicolumn{2}{c|}{Front-end}  & \multirow{2}{*}{Back-end}  \\ 
        & FD+FM\tnote{1} & Others &  \\
        \midrule
        ORB-SLAM3 & \cellcolor{green!20}32.9\second   & \cellcolor{green!20}25.2\second & \cellcolor{green!20}39.6\second     \\
        Ours      & \cellcolor{red!20}17.5\second   & \cellcolor{green!20}8.4\second  & \cellcolor{green!20}49.3\second    \\
        \bottomrule
    \end{tabular}
        \begin{tablenotes}
            \footnotesize
            \item[1] Total runtime for feature detection and feature matching. \textcolor{red!30}{\rule{1em}{0.7em}} and \textcolor{green!30}{\rule{1em}{0.7em}} represent the modules running on the GPU and CPU, respectively.
        \end{tablenotes}
    \end{threeparttable}
    \vspace{-1.5em}
\end{table}

\subsubsection{Mapping Efficiency}

We also evaluate the mapping time, \ie the total runtime for building the initial map and offline optimizing the map. As we compare our system with Hloc using the TartanAir dataset in the map reuse experiment, we use the same baseline and dataset in this experiment. The average mapping time per frame may differ when the map size varies, therefore, we measure the mapping time with different numbers of input images. The results are presented in \fref{fig:mapping_time}, where $n \times$ means our system is $n$ times faster than Hloc. It can be seen that our system is much more efficient than Hloc, especially as the input images increase. Besides, Hloc can only use monocular images to build a map without the real scale, and the map only contains point features, while our system can build the point-line map and estimate the real scale using a stereo camera and an IMU. Therefore, our system is more stable and practical for robotics applications than Hloc.

\subsubsection{Embedded Platform}

\begin{figure}[t]
    \vspace{1.0em}
    \centering
    \setlength{\abovecaptionskip}{-0.01cm}
    \includegraphics[width=0.93\linewidth]{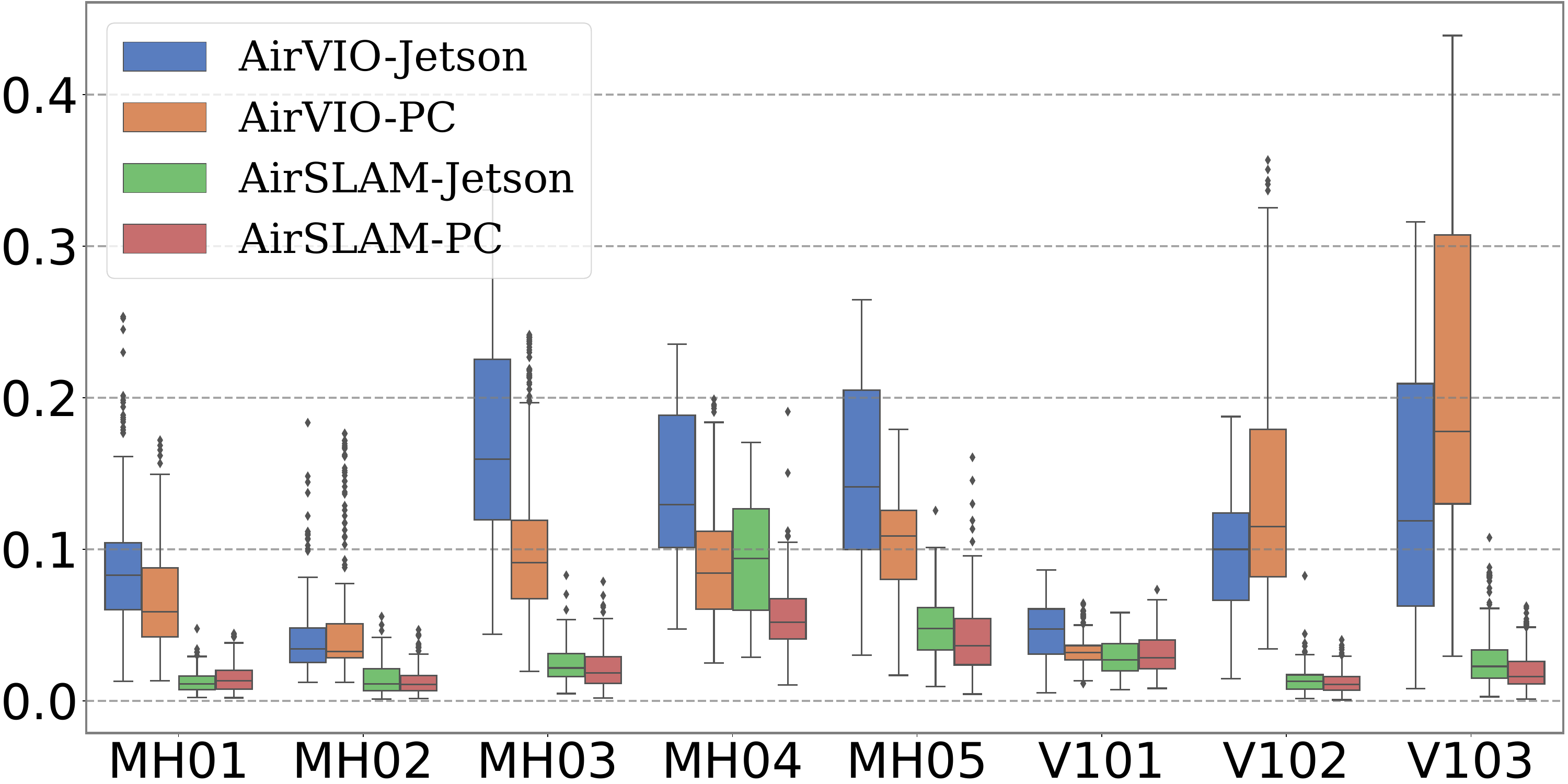}
    \caption{Accuracy comparison of our system on an NVIDIA Jetson Orin and a PC. The vertical axis represents ATE in meters.}
    \label{fig:jetson_pc}
    \vspace{0.2em}
\end{figure}

\begin{table}[t]
    \vspace{0.0em}
    \caption{Efficiency comparison of our system on two platforms.}
    \label{tab:jetson_efficiency}
    \centering  
    \begin{threeparttable}
    \begin{tabular}{C{0.16\linewidth}|C{0.14\linewidth}C{0.14\linewidth}|C{0.14\linewidth}|C{0.16\linewidth}}
        \toprule
        \multirow{2}{*}{Platform} & \multicolumn{2}{c|}{Runtime}  & CPU & GPU  \\ 
        & VIO (FPS) & Optim\tnote{1}. (\second) & Usage (\%) & Usage (MB) \\
        \midrule
        Ours-Jetson & 40.3   & 57.8 & 224.7 & 989    \\
        Ours-PC     & 73.1   & 55.5 & 217.8 & 3076   \\
        \bottomrule
    \end{tabular}
        \begin{tablenotes}
            \footnotesize
            \item[1] The runtime of the offline map optimization.
        \end{tablenotes}
    \end{threeparttable}
    \vspace{-1.5em}
\end{table}

We use 8 sequences in the EuRoC dataset to evaluate the efficiency of AirSLAM on the embedded platform. The suffixes, \ie ``-Jetson" and ``-PC", are added to distinguish results on different platforms. On the Jetson, we modify three parameters in our system to improve efficiency. First, we reduced the number of detected keypoints from 350 to 300. Second, we change two parameters in \sref{sec:kf_selection} to make keyframes sparser, \ie $\alpha_1$ and $\alpha_2$ are changed from 0.65 and 0.1 to 0.5 and 0.2, respectively. The other parameters are the same on these two platforms. 
The comparisons of efficiency and absolute trajectory error (ATE) are presented in \tref{tab:jetson_efficiency} and \fref{fig:jetson_pc}, respectively.
Our AirSLAM can run at a rate of $40\hertz$ on the Jetson while only consuming 2 CPU cores and 989MB GPU memory. We find the runtime of the offline map optimization is very close on these two platforms. This is because AirSLAM-Jetson selects fewer keyframes than AirSLAM-PC, so the loop closure and GBA are faster. 

\section{Conclusion}\label{sec:conclusion}

In this work, we presented an efficient and illumination-robust hybrid vSLAM system. To be robust to challenging illumination, the proposed system employs a CNN to detect both keypoints and structural lines. Then these two features are associated and tracked using a GNN. 
To enhance the efficiency, we proposed PLNet, a unified model capable of simultaneously detecting both point and line features.
Furthermore, a multi-stage relocalization method based on both appearance and geometry information was proposed for efficient map reuse. We designed the system with an architecture that includes online mapping,  offline optimization, and online relocalization, making it easier to deploy on real robots.  
Extensive experiments show that the proposed system outperforms other SOTA vSLAM systems in terms of accuracy, efficiency, and robustness in illumination-challenging environments. 

Despite its remarkable performance, the proposed system still has limitations. Like other point-line-based SLAM systems, our AirSLAM relies on enough line features, so it is best to apply it to man-made environments. This is because our system was originally designed for warehouse robots. In unstructured environments, the system degrades into a point-only system.



\ifCLASSOPTIONcaptionsoff
  \newpage
\fi


{
    \bibliographystyle{IEEEtran}
    \bibliography{papers}
}

{
\normalfont

%

\begin{IEEEbiography}
  [{\includegraphics[width=1in,height=1.25in,clip,keepaspectratio]{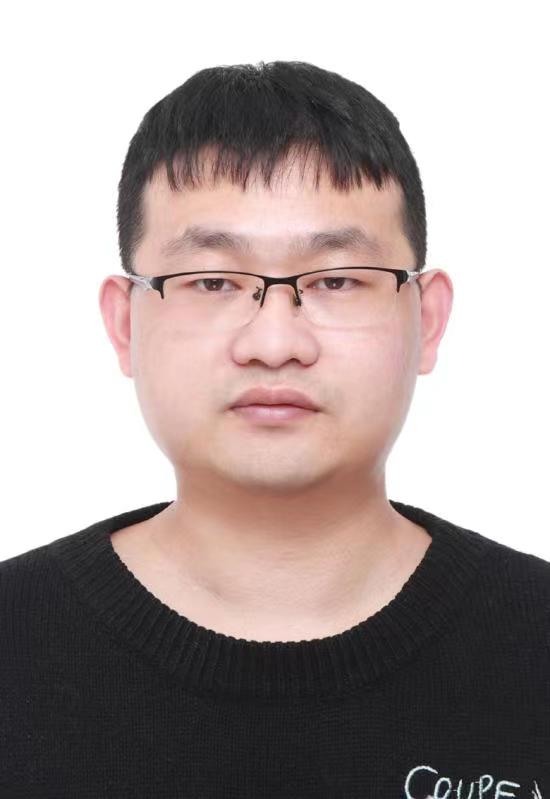}}]
  {Kuan Xu} received the B.E. and M.E. degrees in Electrical Engineering from the Harbin Institute of Technology, Harbin, China, in 2016 and 2018, respectively. He is currently working toward the Ph.D. degree in Electrical and Electronic Engineering with the Department of Electrical and Electronic Engineering, Nanyang Technological University, Singapore.
  From July 2018 to March 2020, he worked as a robot algorithm engineer at Tencent Holdings Ltd, Beijing, China. From March 2020 to April 2022, he served as a senior robot algorithm engineer at Geekplus Technology Co., Ltd., Beijing, China. His research interests include visual SLAM, robot localization, and perception.
\end{IEEEbiography}

\begin{IEEEbiography}
  [{\includegraphics[width=1in,height=1.25in,clip,keepaspectratio]{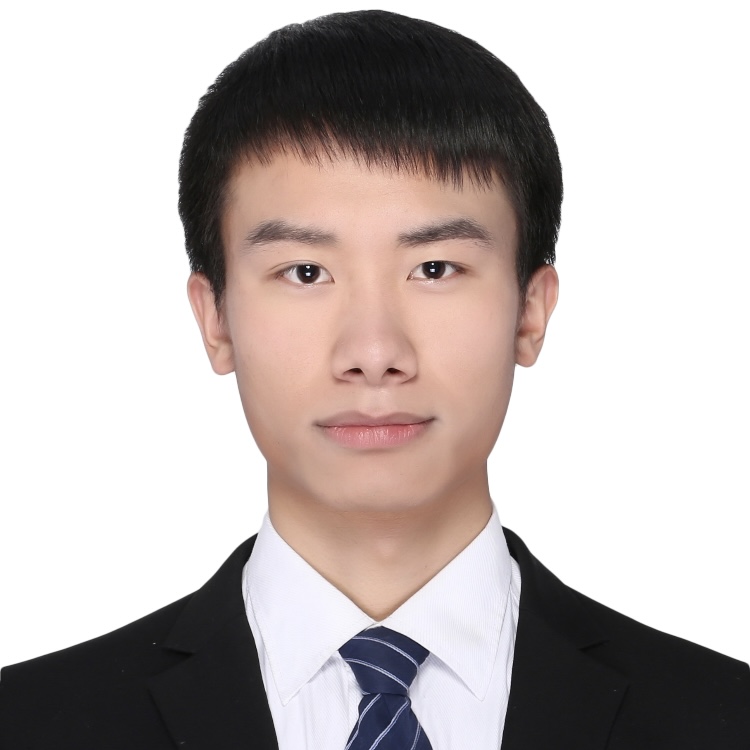}}]
  {Yuefan Hao} received a B.E. degree in Communication Engineering from Kunming University of Science and Technology, Kunming, China, in 2017 and a M.E. degree in Electrical and Communication Engineering from University of Electronic Science and Technology of China, Chengdu, China, in 2020. From June 2020 to June 2024, he served as a robot algorithm engineer at Geekplus Technology Co., Ltd., Beijing, China. His research interests include computer vision, deep learning, and robotics. 
\end{IEEEbiography}

\begin{IEEEbiography}
  [{\includegraphics[width=1in,height=1.25in,clip,keepaspectratio]{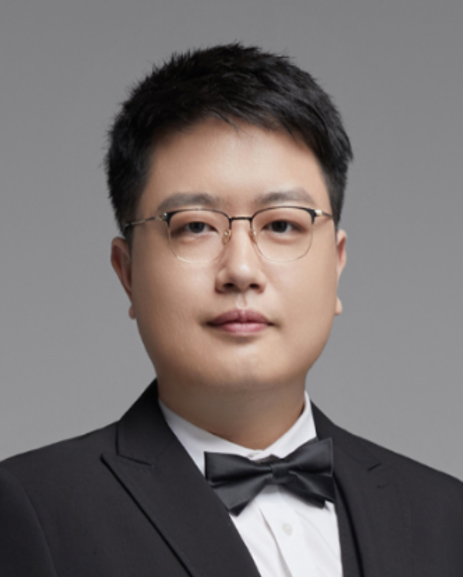}}]
  {Shenghai Yuan} (Member, IEEE) received his B.S. and Ph.D. degrees in Electrical and Electronic Engineering in 2013 and 2019, respectively, from Nanyang Technological University, Singapore. His research focuses on robotics perception and navigation. He is a postdoctoral senior research fellow at the Centre for Advanced Robotics Technology Innovation (CARTIN) at Nanyang Technological University, Singapore. He has contributed over 70 papers to journals such as TRO, IJRR, TIE, and RAL, and to conferences including ICRA, CVPR, ICCV, NeurIPS, and IROS. He has also contributed over 10 technical disclosures and patents. Currently, he serves as an associate editor for the Unmanned Systems Journal and as a guest editor for the Electronics Special Issue on Advanced Technologies of Navigation for Intelligent Vehicles. He achieved second place in the academic track of the 2021 Hilti SLAM Challenge, third place in the visual-inertial track of the 2023 ICCV SLAM Challenge, and won the IROS 2023 Best Entertainment and Amusement Paper Award. He also received the Outstanding Reviewer Award at ICRA 2024. He served as the organizer of the CARIC UAV Swarm Challenge and Workshop at the 2023 CDC and the UG2 Anti-drone Challenge and Workshop at CVPR 2024. Currently, he is the organizer of the second CARIC UAV Swarm Challenge and Workshop at IROS 2024. 
\end{IEEEbiography}



\begin{IEEEbiography}[{\includegraphics[width=1in,height=1.25in,clip,keepaspectratio]{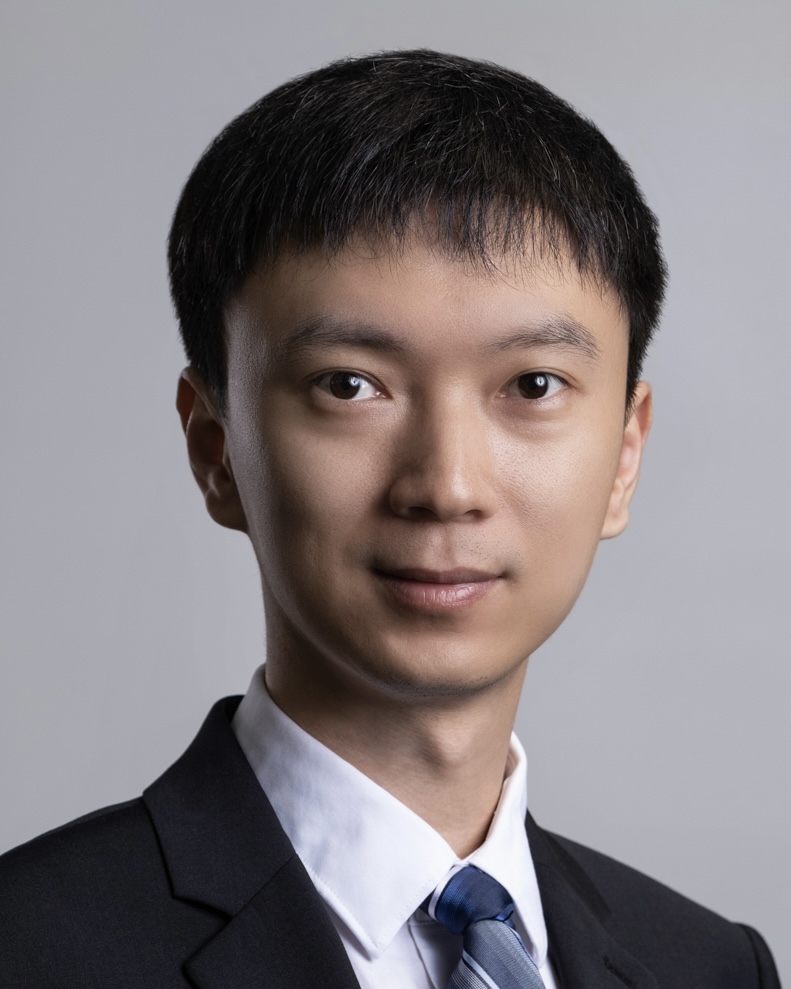}}]
{Chen Wang} (Senior Member, IEEE) received a B.Eng. degree in Electrical Engineering from Beijing Institute of Technology (BIT) in 2014 and a Ph.D. degree in Electrical Engineering from Nanyang Technological University (NTU) Singapore in 2019. He was a Postdoctoral Fellow with the Robotics Institute at Carnegie Mellon University (CMU).

Dr. Wang is an Assistant Professor and leading the Spatial AI \& Robotics (SAIR) Lab at the Department of Computer Science and Engineering, University at Buffalo (UB). 
He is an Associate Editor for the International Journal of Robotics Research (IJRR) and IEEE Robotics and Automation Letters (RA-L) and an Associate Co-chair for the IEEE RAS Technical Committee (TC) for Computer \& Robot Vision. He served as an Area Chair for the IEEE/CVF Conference on Computer Vision and Pattern Recognition (CVPR), the IEEE International Conference on Robotics and Automation (ICRA), and the Conference on Neural Information Processing Systems (NeurIPS). His research interests include Spatial AI and Robotics.
\end{IEEEbiography}

\begin{IEEEbiography}
  [{\includegraphics[width=1in,height=1.25in,clip,keepaspectratio]{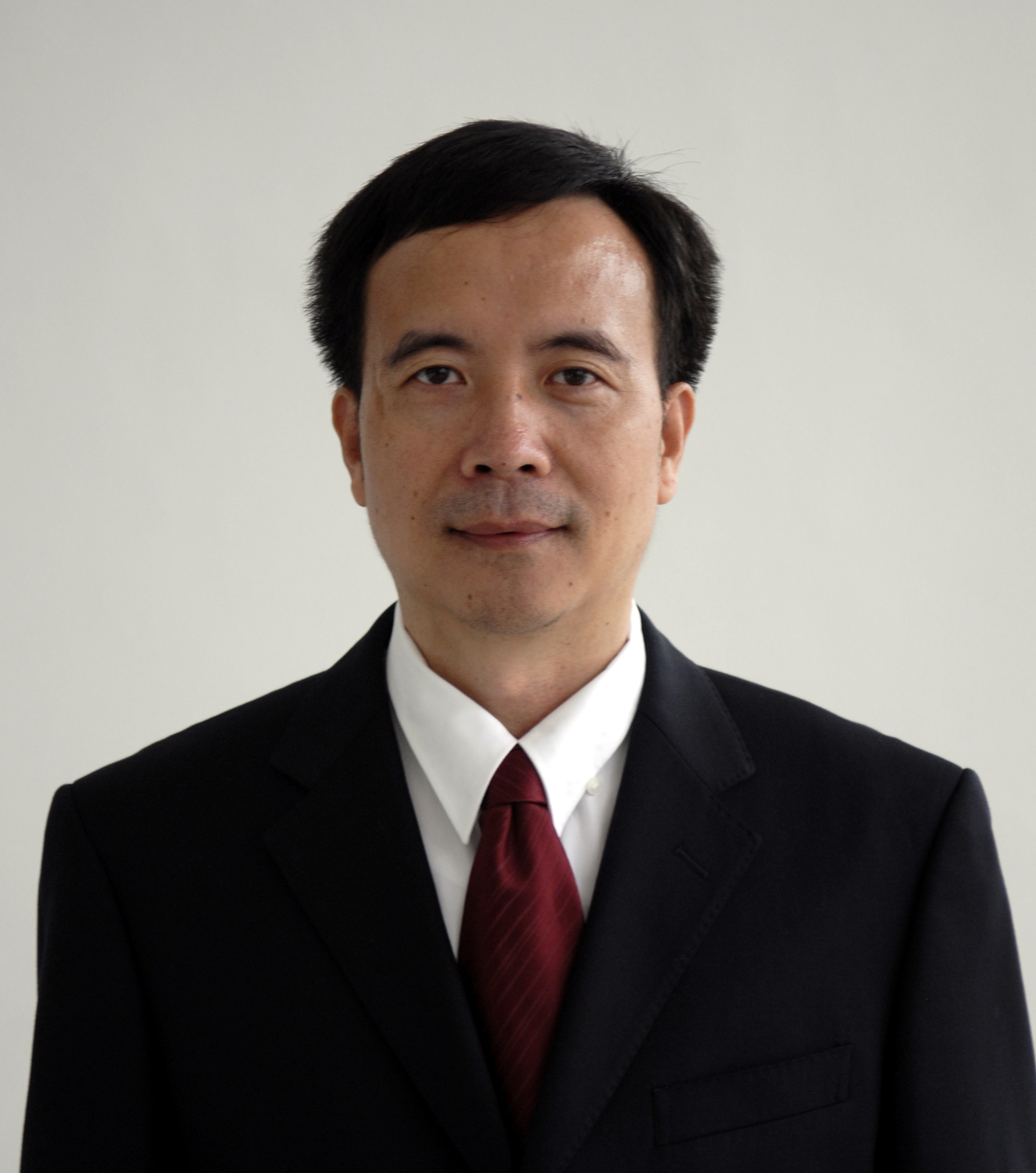}}]
  {Lihua Xie} (Fellow, IEEE) received the Ph.D. degree in electrical engineering from the University of Newcastle, Australia, in 1992. Since 1992, he has been with the School of Electrical and Electronic Engineering, Nanyang Technological University, Singapore, where he is currently a President's Chair and Director, Center for Advanced Robotics Technology Innovation. He served as the Head of Division of Control and Instrumentation and Co-Director, Delta-NTU Corporate Lab for Cyber-Physical Systems. He held teaching appointments in the Department of Automatic Control, Nanjing University of Science and Technology from 1986 to 1989. 
  Dr Xie’s research interests include robust control and estimation, networked control systems, multi-agent networks, and unmanned systems. He is an Editor-in-Chief for Unmanned Systems and has served as Editor of IET Book Series in Control and Associate Editor of a number of journals including IEEE Transactions on Automatic Control, Automatica, IEEE Transactions on Control Systems Technology, IEEE Transactions on Network Control Systems, and IEEE Transactions on Circuits and Systems-II. He was an IEEE Distinguished Lecturer (Jan 2012 – Dec 2014). Dr Xie is Fellow of Academy of Engineering Singapore, IEEE, IFAC, and CAA.
\end{IEEEbiography}

}

\vfill


\end{document}